\documentclass[final,3p,times,twocolumn]{elsarticle}
\usepackage{my_style}
\usepackage{lineno, hyperref}
\modulolinenumbers[5]

\journal{Journal of Image and Vision Computing}

\begin{document}

\begin{frontmatter}

\title{RoI Tanh-polar Transformer Network for Face Parsing in the Wild}

\author{Yiming Lin}
\ead{yiming.lin15@imperial.ac.uk}
\author{Jie Shen\corref{mycorrespondingauthor}}
\cortext[mycorrespondingauthor]{Corresponding author}
\ead{jie.shen07@imperial.ac.uk}
\author{Yujiang Wang}
\ead{yujiang.wang14@imperial.ac.uk}
\author{Maja Pantic}
\ead{maja.pantic@gmail.com}

\address{Department of Computing, Imperial College London, UK}

\begin{abstract}
Face parsing aims to predict pixel-wise labels for facial components of a target face in an image.
Existing approaches usually crop the target face from the input image with respect to a bounding box calculated during pre-processing, and thus can only parse inner facial Regions of Interest~(RoIs). Peripheral regions like hair are ignored and nearby faces that are partially included in the bounding box can cause distractions. Moreover, these methods are only trained and evaluated on near-frontal portrait images and thus their performance for in-the-wild cases has been unexplored.
To address these issues, this paper makes three contributions. First, we introduce iBugMask dataset for face parsing in the wild, which consists of $21,866$ training images and $1,000$ testing images. The training images are obtained by augmenting an existing dataset with large face poses. The testing images are manually annotated with $11$ facial regions and there are large variations in sizes, poses, expressions and background.
Second, we propose RoI Tanh-polar transform that warps the whole image to a Tanh-polar representation with a fixed ratio between the face area and the context, guided by the target bounding box. 
The new representation contains all information in the original image, and allows for rotation equivariance in the convolutional neural networks~(CNNs).
Third, we propose a hybrid residual representation learning block, coined HybridBlock, that contains convolutional layers in both the Tanh-polar space and the Tanh-Cartesian space, allowing for receptive fields of different shapes in CNNs. Through extensive experiments, we show that the proposed method improves the state-of-the-art for face parsing in the wild and does not require facial landmarks for alignment. 

\end{abstract}


\end{frontmatter}

\section{Introduction}
\label{sec:intro}

Face parsing is a fundamental facial analysis task: it predicts per-pixel semantic labels in a target face. It provides useful features for many downstream applications, such as face recognition~\cite{chen2019matching, Cheng2019}, face beautification \cite{ou2016beauty}, face swapping \cite{Nirkin2018, Nirkin2019}, face synthesis \cite{Banerjee2020, li2017generative, zhang2019synthesis, Chen_2020_CVPR, Zhu_2020_CVPR}, facial attribute recognition~\cite{kalayeh2017improving, kalayeh2019symbiosis,khan2020multi}, and facial medical analysis~\cite{TCMINet}. Recently, methods based on deep Convolutional Neural Networks (CNNs), especially Fully Convolutional Networks (FCNs)~\cite{FCNs}, have achieved impressive results on this task.\cite{Lin_2019_CVPR, luo2012hier, liu2015multi, masi2020structure, Wei2019AccurateFI, jackson2016cnn, guo2018res}. 

Two unique aspects distinguish face parsing from generic image parsing. The first is that face parsing does not parse the entire image but only the target face specified by a bounding box, whereas image parsing predicts a label for every pixel in the image. 
How to preprocess the input with bounding boxes remains an under-explored problem. Most methods~\cite{wei2017learning,Liu2017Face} crop the face with a fixed margin and resize the cropped patch to the same dimension, which we refer to as \emph{crop-and-resize} the face area.
Such methods ignore hair because the margin around the hair area is hard to determine. If the selected margin is too small, the hair region would be cut off. If it is too big, too many background pixels and / or nearby faces could be included in the cropped patch, causing significant distractions to the model. Another pre-processing method is to use facial landmarks for face alignment~\cite{Lin_2019_CVPR} such that the face is appropriately rotated. We refer to this method as \emph{align}. The landmarks can be jointly obtained with the face bounding boxes~\cite{retinaface}. The main problem is that a good template has to be carefully chosen for alignment.

\begin{figure*}[t]
    \begin{center}
        
    \subfloat[CelebAMask-HQ]{{\includegraphics[width=0.19\linewidth, height=4.8cm]{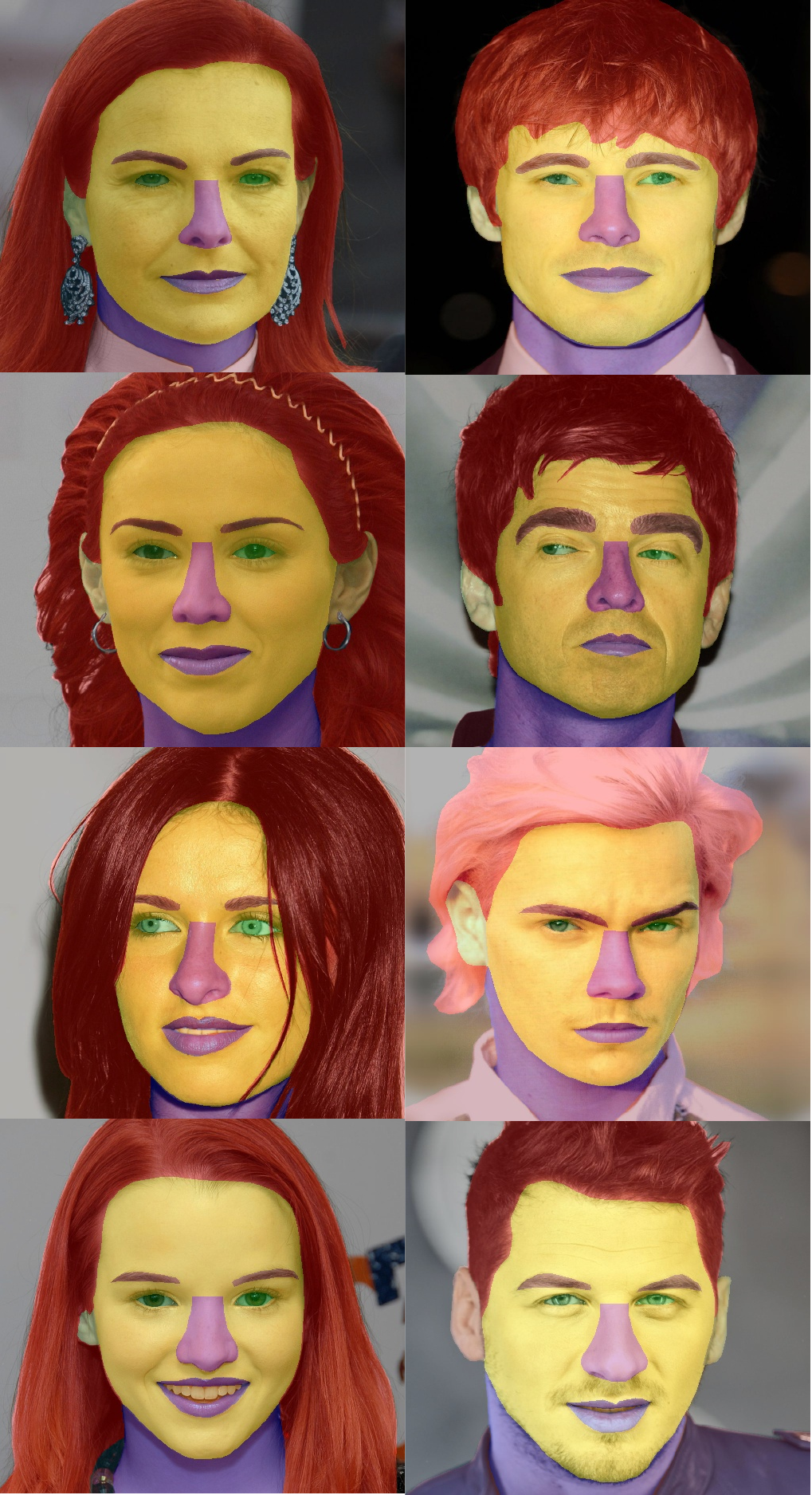} }}%
    \subfloat[LFW-PL]{{\includegraphics[width=0.18\linewidth, height=4.8cm]{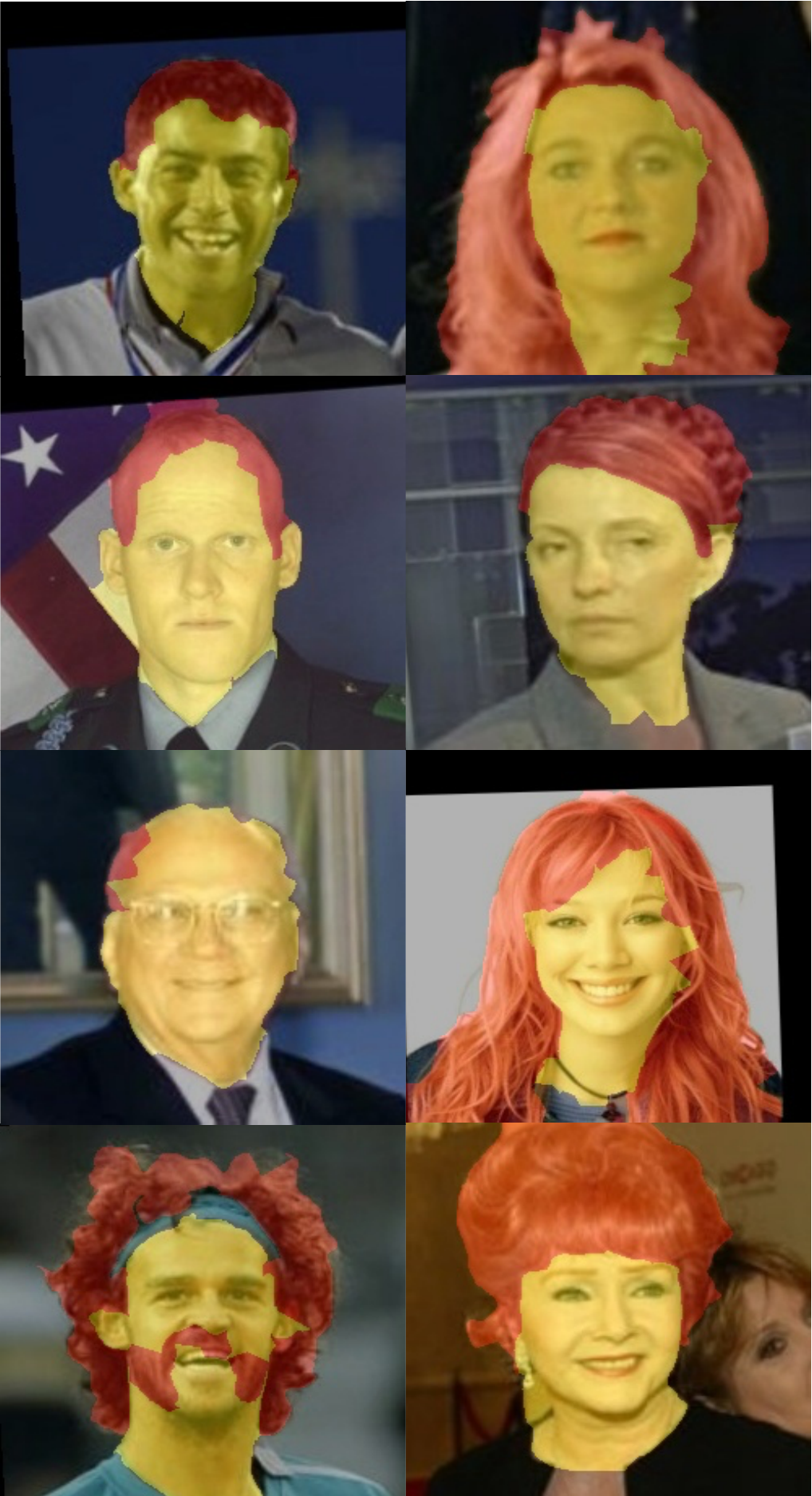} }}%
    \subfloat[Helen]{{\includegraphics[width=0.18\linewidth, height=4.8cm]{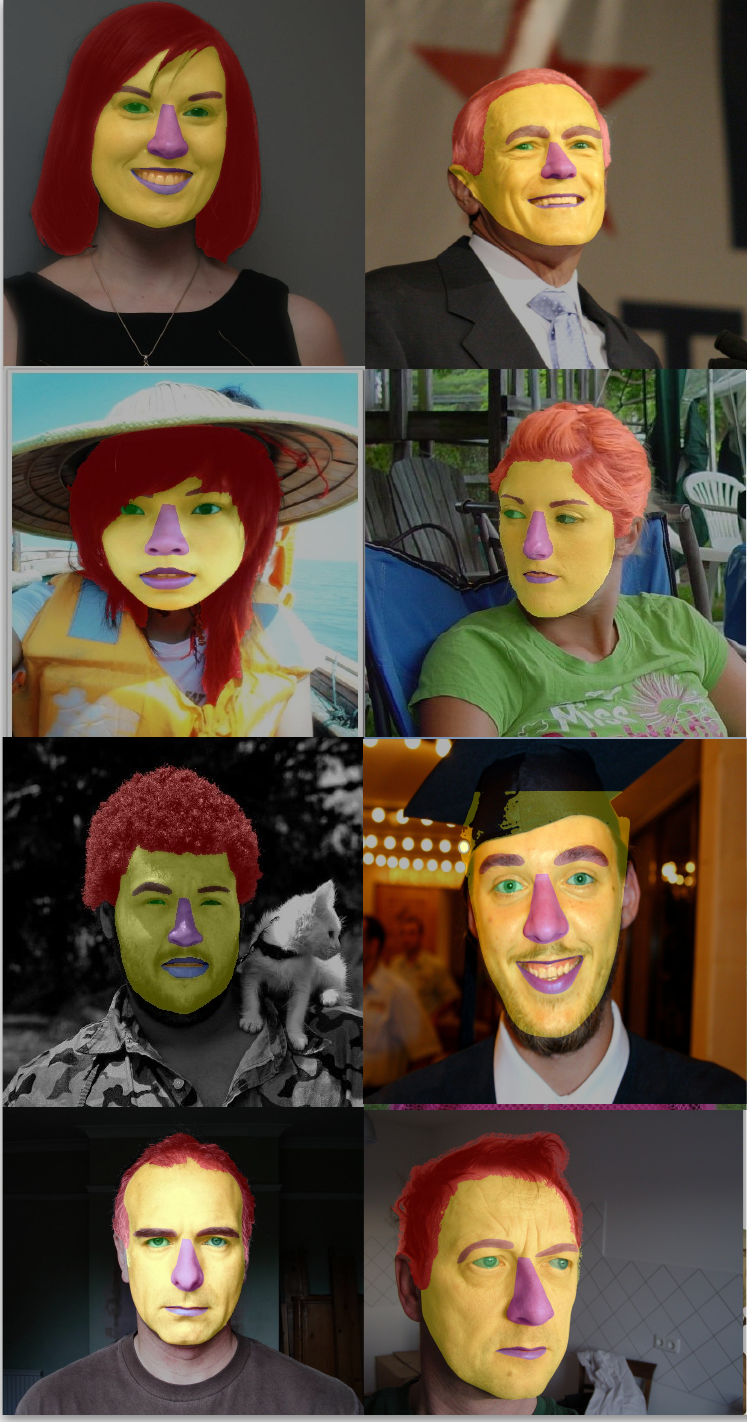} }}%
    \subfloat[LaPa]{{\includegraphics[width=0.18\linewidth, height=4.8cm]{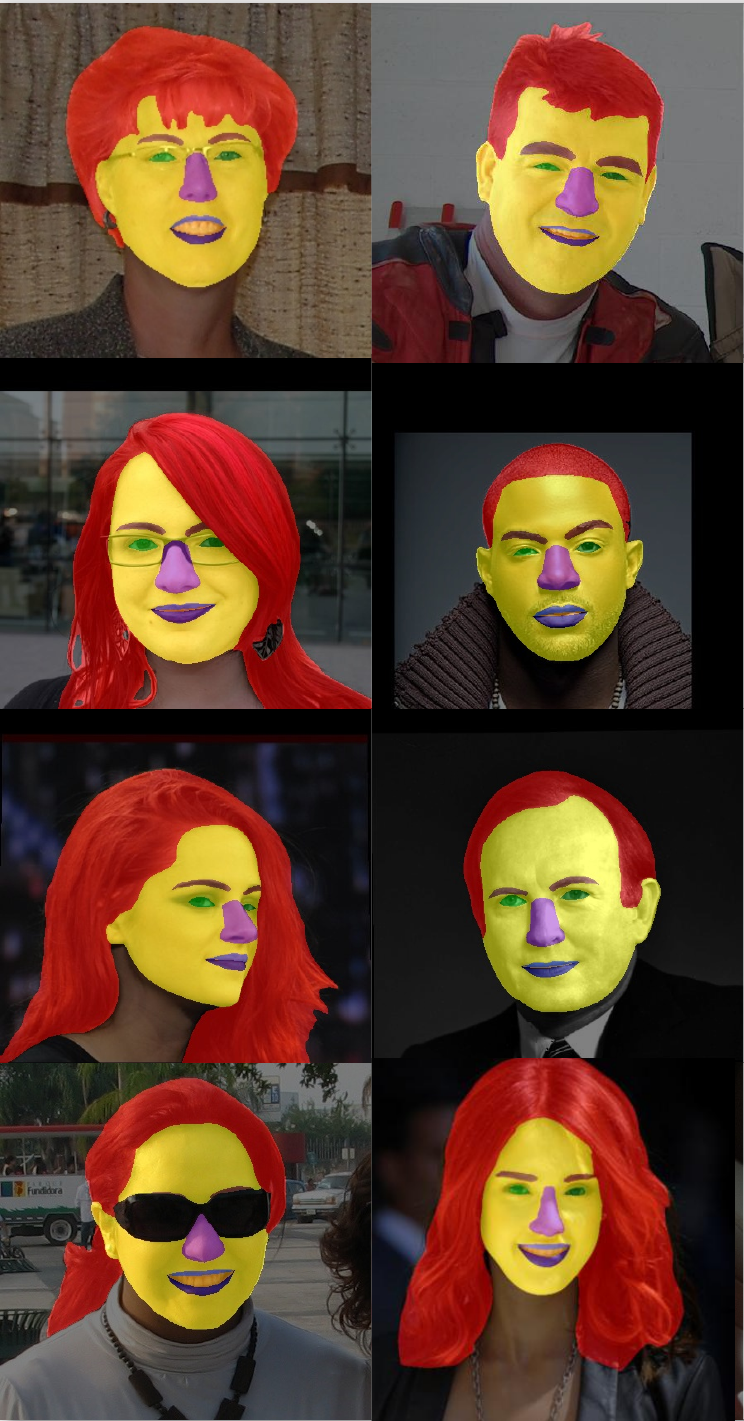} }}%
    \subfloat[iBugMask]{{\includegraphics[width=0.18\linewidth, height=4.8cm]{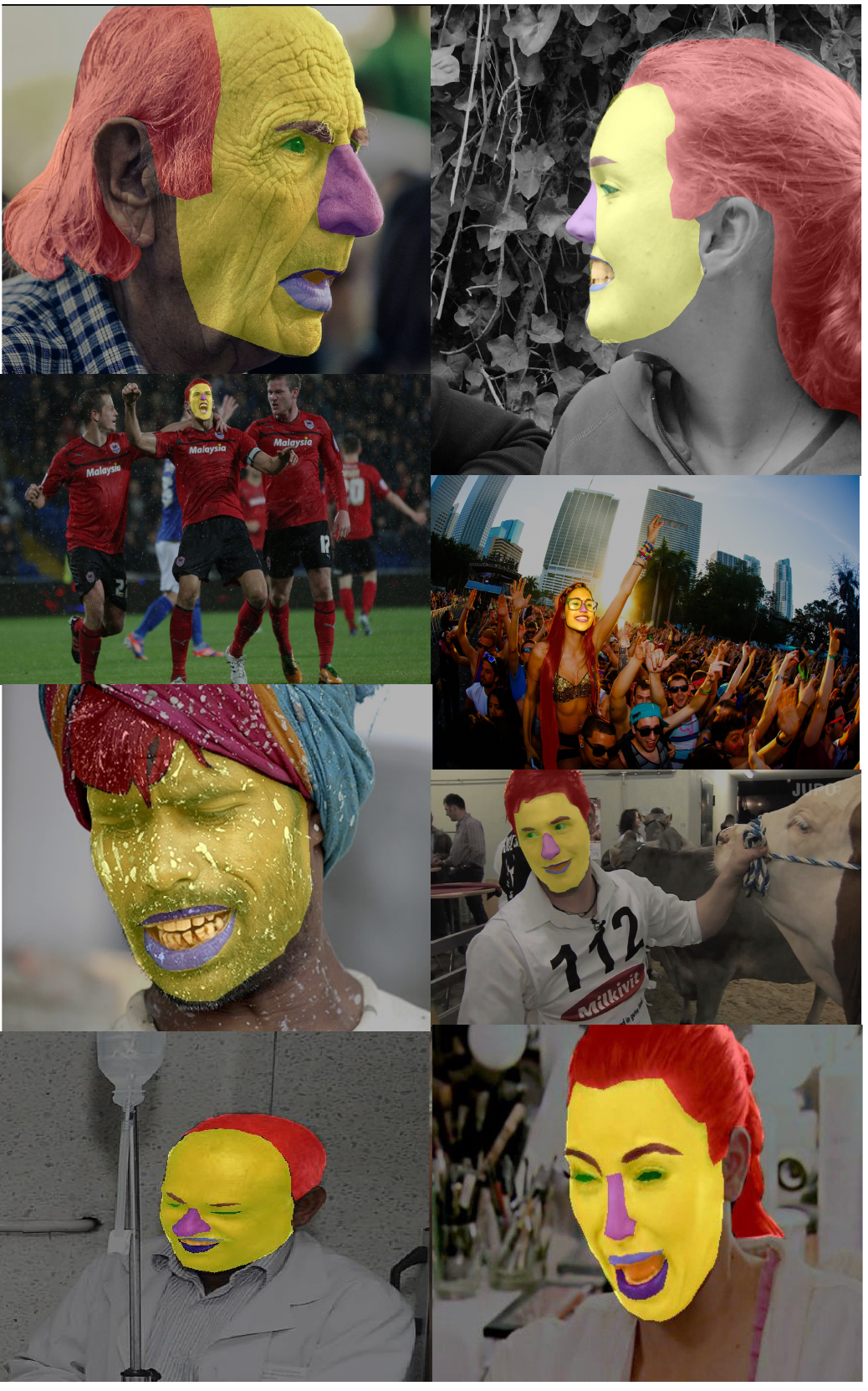} }}%
    \caption{Examples from benchmarks with colour-coded labels (best viewed in colour). CelebAMask-HQ~\cite{CelebAMask-HQ} and LFW-PL~\cite{lfwpl} contain well-aligned faces and little context information. Helen~\cite{Smith_2013_CVPR} contains mostly portrait images where faces are big and near the centre. LaPa~\cite{lapa} contains face images with some variations in pose and occlusion but the faces are cropped and centred. 
    By contrast, iBugMask contains large variations in expression, pose and background and all background information is preserved (see Section \ref{sec:data}).}%
    \label{fig:benchmarks}%
\end{center}

\end{figure*}

The other challenge is that in-the-wild images are underrepresented in existing benchmarks.
Four most widely used face parsing datasets are Helen~\cite{Smith_2013_CVPR}, LFW-PL~\cite{lfwpl}, CelebAMask-HQ~\cite{CelebAMask-HQ} and LaPa~\cite{lapa}. \figurename~\ref{fig:benchmarks} shows exemple images from these datasets and Section~\ref{sec:data} compares them in detail.
Most images from Helen and LaPa are portraits, which means that only one large face in frontal view is present near the centre. 
CelebAMask-HQ contains images synthesised from CelebA~\cite{liu2015faceattributes} using super-resolution. All faces are aliened using landmarks and resized to the same size. As such, the resulted images contain very little context information. Similarly, LFW-PL is a subset of LFW~\cite{LFWTech} and the faces are aligned using landmarks. Hence, how models perform under in-the-wild conditions remains unexplored. 

To tackle the first challenge, we propose RoI Tanh-polar transform (RT-Transform), that transforms the entire image into a fixed-size representation in the Tanh-polar coordinate system based on the target bounding box.  \figurename~\ref{fig:roi-tp-tsfm} illustrates the transform process. 
As a fully invertible transform, it preserves all contextual information. Moreover, regardless of the face's actual size in the input image, the ratio between the face and the background remains fixed at $76\%:24\%$ in the transformed representation. In the Tanh-polar coordinate system, planar convolutions correspond to group-convolutions~\cite{cohen2016group} in rotation. Thus, Convolutional Neural Networks~(CNNs) applied in the tanh-polar space would produce a representation that is equivariant to rotations in the original Cartesian space.

\begin{figure*}[!tb]
    \centering
    \includegraphics[width=\textwidth]{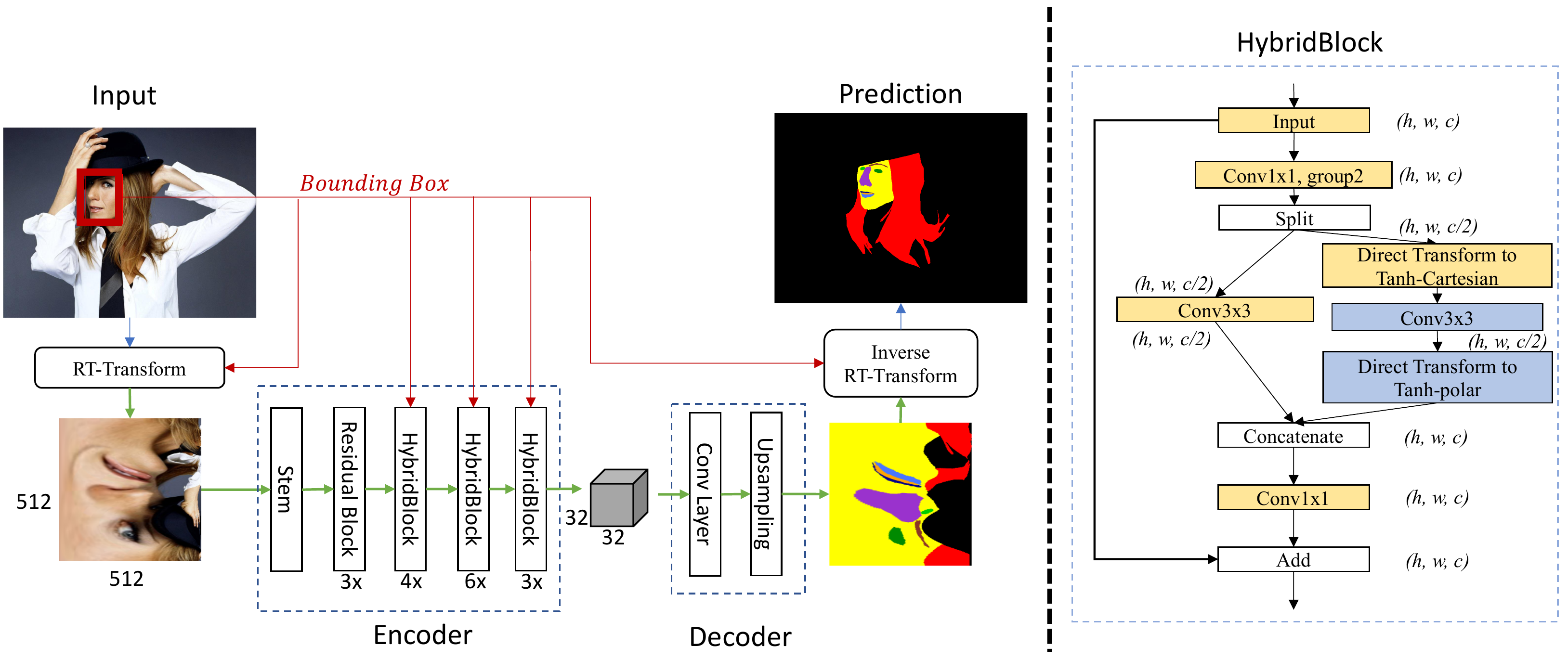}
    \caption{\emph{Left:} RoI Tanh-polar Transformer Network (RTNet): a facial image is transformed to Tanh-polar coordinates with RT-Transform. The encoder consists of a Stem layer, one stage of Residual Blocks and three stages of HybridBlocks. Bounding box is used in RT-transform and HybridBlocks to warp tensors between Tanh-polar and Tanh-Cartesian coordinates.
    The decoder consists of a Conv layer and a Bilinear Upsampling layer. The output mask is transformed back to Cartesian coordinates using inverse RT-Transform. \emph{Right:}  HybridBlock. Yellow rectangles are layers in Tanh-polar space while blue ones are layers in Tanh-Cartesian space. Tuples $(h,w,c)$ are the shape of the output tensor for each operation. ``Split'' and ``Concatenate'' operations are performed along the channel dimension (see Section~\ref{sec:HybridBlock}).}
    \label{fig:overall}
\end{figure*}

We further introduce Hybrid Residual Representation Learning Block (HybridBlock) that uses RT-Transform to create hybrid representations in the residual blocks. A HybridBlock consists of two $3\times 3$ convolutional layers, one in the Tanh-polar coordinate system and the other in the Tanh-Cartesian coordinate system.  
They are operating on different-shaped receptive fields and their outputs are concatenated in the Tanh-polar system to obtain a hybrid representation. By stacking HybridBlocks, we arrive at HybridNet, a backbone network that takes as input an image transformed by RT-Transform and the target bounding box, and outputs a hybrid representation for face parsing. We then add the vanilla FCN decoder and inverse RT-Transform. The resulting framework, called RoI Tanh-polar Transformer Network~(RTNet), is shown in \figurename~\ref{fig:overall}.

To tackle the second challenge of lacking suitable benchmarks, we present iBugMask dataset that contains $22,866$ in-the-wild images. For the training set of $21,866$ images, we use a face profiling method~\cite{3ddfa} to rotate the faces from images in Helen dataset with respect to the yaw angle, creating many large-pose and profile faces. For the $1,000$ testing images, per-pixel manual annotations for $11$ regions including hair are provided. The curated images contain large variations in pose, expression, size and background clutter (see \figurename~\ref{fig:benchmarks}). 
Extensive experiments show that iBugMask dataset is more challenging than other benchmarks and models trained on iBugMask improves performance under both intra-dataset and cross-dataset evaluation.

In summary, we offer the following contributions:
\begin{itemize}
    \item We propose RoI Tanh-polar Transform for face parsing in the wild that transforms the target face to the Tanh-polar coordinate system based on the bounding box, preserving the context and allowing CNNs to learn representations equivariant to rotations.
    \item We propose Hybrid Residual Representation Learning Blocks, that extracts a hybrid representation by applying convolutions in both Tanh-polar and Tanh-Cartesian coordinates.
    \item We present iBugMask dataset, a novel in-the-wild face parsing benchmark that consists of more than $22$ thousand images.
    \item We conduct extensive experiments and show that the overall framework RTNet improves the state-of-the-art on all benchmarks.
\end{itemize}
\section{Related Work}
\label{sec:related-word}
Increasing research effort has been devoted to face parsing due to its potential application in various face analysis tasks. In this section, we briefly review four groups of relevant works, \ie~1) the face parsing benchmarks, 2) the face parsing methods, 3) works on scene parsing and 4) representation learning in polar space.

\subsection{Face Parsing Benchmarks}\label{sec:face-parsing-datasets}
Publicly available face parsing benchmarks are comparatively scarce, mainly due to the significant amount of effort required for pixel-level annotations. Currently, two most widely-used benchmarks are LFW-PL~\cite{lfwpl} and Helen~\cite{Smith_2013_CVPR}. 

The Helen dataset~\cite{le2012interactive,Smith_2013_CVPR} includes $2,330$ facial images with 194 landmark annotations that are obtained through Amazon Mechanical Truck. In this dataset, the inner facial components including eyes, eyebrows, nose, inner mouth and upper/lower lips are manually annotated by human, while the ground-truths for the rest facial parts, \ie~facial skin and hairs, are generated via image matting algorithms \cite{Smith_2013_CVPR} and may not be fully accurate. Despite such disadvantages, Helen was still the only publicly-available face parsing dataset with an acceptable amount of training data for several years, and thus it has been a popular choice for evaluating face parsing methods~\cite{Lin_2019_CVPR,Liu2017Face,te2020edge}.

The LFW-PL dataset~\cite{lfwpl} consists of $2,972$ facial images selected from the Labeled Faces in the Wild (LFW) dataset~\cite{LFWTech}. 
To obtain dense annotations, each facial image is first automatically segmented into superpixels and those superpixels are subsequently labelled as one of the following categories: facial skin, hair and background. 

Recently, two large-scale face parsing datasets were released, which are CelebAMask-HQ~\cite{CelebAMask-HQ} and LaPa~\cite{lapa}. Although the number of annotated samples are greatly increased, the facial images included in those two datasets are not strictly in-the-wild, since they have already been pre-processed in an unrecoverable way.
In CelebAMask-HQ, the resolutions of facial images are intentionally enlarged through the super-resolution technique \cite{karras2017progressive}, while most faces are aligned to be frontal and centralised. Besides, the background region usually comprises a small portion of the whole facial image, \ie~most environmental information has been discarded. Similar situations can be discovered in the Lapa dataset in which faces are also cropped and aligned with limited background information preserved.

Compared to those datasets, our proposed iBugMask dataset is the only face parsing benchmark consisting of fully in-the-wild images. The facial samples are neither cropped nor aligned, and we also preserve almost all the background information. It covers large variations in poses, illuminations, occlusions and scenes. A detailed comparison between the iBugMask dataset and the existing benchmarks is provided in Sec.~\ref{sec:data}.

\subsection{Face Parsing Methods}
Face parsing is the task of pixel-wisely labelling given facial images. 
Earlier works \cite{Warrell2009,Smith_2013_CVPR} on face parsing usually leveraged holistic priors and hand-crafted features. Warrell~\etal~\cite{Warrell2009} modelled the spatial correlations of facial parts with Conditional Random Fields~(CRFs). Smith~\etal~\cite{Smith_2013_CVPR} applied SIFT features to select exemplars in facial parts and propagate the labels of these exemplars to generate complete segmentation maps. A hybrid method was proposed in~\cite{lfwpl} that combined the strength of both CRFs and Restricted Boltzmann Machine in a single framework to model global and local facial structures. 
The idea of utilising engineering-based features can also been seen in other works \cite{scheffler2011joint,yacoob2006detection,lee2008markov}. Those approaches are typically time-consuming and cannot generalise well to different scenarios, and thus they have been gradually replaced by deep-learning-based methods with encouraging performance.  

State-of-the-art performance on face parsing is mostly achieved by deep learning methods. 
Liu~\etal~\cite{liu2015multi} incorporated CNNs into CRFs and proposed a multi-objective learning method to model pixel-wise likelihoods and label dependencies jointly. 
An interlinked CNN was present in \cite{zhou2015interlinked} to detect different facial parts, while this architecture cannot generate semantic labels for large-scale components like facial skin.
Luo~\etal~\cite{luo2012hier} applied multiple Deep Belief Networks to detect facial parts and accordingly built a hierarchical face parsing framework. Jackson~\etal~\cite{jackson2016cnn} employed facial landmarks as a shape constraint to guide Fully Convolution Networks~(FCNs) for face parsing. 
Multiple deep methods including CRFs, Recurrent Neural Networks (RNNs) and Generative Adversarial Networks (GAN) were integrated by authors of \cite{gucclu2017end} to formulate an end-to-end trainable face parsing model, while the facial landmarks also served as the shape constraints for segmentation predictions. The idea of leveraging shape priors to regularise segmentation masks can also be found in the Shape Constrained Network (SCN) \cite{luo2020shape} for eye segmentation.
In~\cite{Liu2017Face}, a spatial Recurrent Neural Networks was used to model the spatial relations within face segmentation masks. 
A spatial consensus learning technique was explored in \cite{masi2020structure} to model the relations between output pixels, while graph models was adopted in \cite{te2020edge} to learn the implicit relationships between facial components. 
To better utilise the temporal information of sequential data, authors of \cite{wang2019face} integrated ConvLSTM \cite{xingjian2015convolutional} with the FCN model \cite{FCNs} to simultaneously learn the spatial-temporal information in face videos and to obtain temporally-smoothed face masks. 
In \cite{Wang_2020_CVPR}, a Reinforcement-Learning-based key scheduler was introduced to select online key frames for video face segmentation such that the overall efficiency can be globally optimised.

Most of those methods assume the target face has already been cropped out and are well aligned. Moreover, they often ignore the hair class due to the unpredictable margins for cropping the hair region. The most related work to our paper is the RoI Tanh Warping~\cite{Lin_2019_CVPR} which proposed to warp the entire image using the Tanh function. However, there are several limitations in this work.
The warping operation requires not only the facial bounding boxes but also the facial landmarks, 
which can be overly redundant. 
Additionally, the warped image is still in Cartesian space 
and cannot benefit from the rotation-equivariant property in polar space. 
Moreover, multiple sub-networks are employed to learned the shapes of inner facial parts and hair separately, and these sub-networks need to be trained with different loss functions, making the pipeline trivial.
\subsection{Scene Parsing}
Scene parsing is to segment an image into different image regions associated with semantic scene labels such as roads, pedestrians, cars, etc.  
Fully Convolutional Networks (FCNs)~\cite{FCNs} is the first critical milestone of applying deep learning techniques in this field. Via replacing fully connected layers with convolutional ones, FCNs successfully adapt classical CNN classification models like VGG-16 \cite{simonyan2014very} or ResNet \cite{resnet} to solve scene parsing tasks. 
Following FCNs~\cite{FCNs}, a wide variety of scene parsing methods have been developed, including the application of dilated (\textit{atrous}) convolutions \cite{yu2015multi,yu2017dilated,deeplab,wang2020dilated}, the encoder and decoder structures \cite{segnet, deeplabv3plus,lin2017refinenet}, the spatial pyramid architectures \cite{pspnet,deeplabv3plus,spnet}
the involvement of attention mechanisms \cite{ccnet,DANet,zhu2019asymmetric}, utilising Neural Architecture Search (NAS) techniques \cite{liu2019auto,dpc2018}, etc.
PSPNet~\cite{pspnet} proposed a spatial pyramid pooling module that adopts a set of spatial pooling operations of different sizes to increase the variety of receptive fields of the network. SPNet~\cite{spnet} extended the pooling module by introducing the strip pooling module to capture long-range banded context. Deeplab family~\cite{deeplab, deeplabv3plus} devised an Atrous Spatial Pyramid Pooling (ASPP) module that consists of three parallel dilated convolutional layers to capture multi-scale context.
UNet~\cite{unet} introduced skip connections between the encoder and the decoder sub-networks to preserve low-level details, while the details of high resolution features were maintained in HRNet~\cite{hrnet} by branching the backbone network. BiSeNet~\cite{bisenet} proposed a bilateral network consisting of a context branch and a spatial branch. Such a two-branch architecture allows BiSeNet to operate with satisfying efficiency while achieving the state-of-the-art performance. Readers are referred to \cite{minaee2020image} for a more detailed review of scene parsing techniques.

\subsection{Polar Representation Learning}
Compared with Cartesian coordinate system, polar or log-polar space are not sensitive to certain transformations such as rotations and scaling, and therefore polar representations have been widely studied in image processing and computer vision.
Early applications of polar transformations included face detection~\cite{hotta_scale_1998}, face tracking~\cite{Jurie1999}, face recognition~\cite{bao_facial_2011}, the aggregation of hand-crafted descriptors \cite{mikolajczyk2005performance,tola2008fast}, etc. \\
Recently, how to integrate polar representations with deep CNN models have been increasingly explored. The traditional CNN architectures can be insensitive to translations, i.e. translation equivalence, yet it is not the case for other transformations such as rotations and scaling. Representation learning in polar space, on the other hand, can effectively overcome such limitations through its equivariances to rotations and scales.  
Polar Transformer Networks (PTN)~\cite{esteves2018polar} is one of the first attempts to construct a CNN model that maps Cartesian-based images into polar coordinates for better tolerances to transformations like rotations and dilation. 
In PTN, a shallow network consisting of several $1\times 1$ convolutional layers first scans the whole image to predict a polar origin. This predicted origin together with the input image are then fed into a differentiable polar transformer module to generate image representations in log-polar systems.  
The obtained polar representation is invariant with respect to the original object locations while rotations and dilations are now shifts, which are handled equivariantly by a conventional classifier CNN. Ebel~\etal~\cite{Ebel_iccv2019_beyond} utilises PTN to extract polar-based local descriptors for key-point matching, leading to more robust performance. 
Different from those works, our RoI Tanh-polar Transformer network warps the whole image into a Tanh-polar representation that can emphasise the Region of Interests (RoI) through oversampling in RoI areas and undersampling in the rest. 
\section{Dataset}
\label{sec:data}

In this section, we introduce a new in-the-wild face parsing benchmark, iBugMask, that consists of a pose-augmented training set and a manully-annotated testing set. We compare their characteristics with existing face parsing datasets. The main motivation for the new benchmark is that existing benchmarks only contain faces with limited variations in expression, pose and context information, which makes them less suitable for capturing characteristics of real-world face images. Moreover, large scale training data is key to the success of CNN-based models, but existing face parsing datasets do not provide sufficient training data for such challenging cases.

\subsection{Overview of Existing Benchmarks}
\figurename~\ref{fig:benchmarks} shows exemplar images from different face parsing benchmarks with their colour-coded labels overlaid.

\textbf{CelebAMask-HQ}~\cite{CelebAMask-HQ} contains $30,000$ synthesised faces from the CelebA dataset~\cite{liu2015faceattributes}. All images are scaled to $512\times512$ and faces are well-aligned at the centre using facial landmarks. Background information is either removed or blurred. $19$ facial classes are labelled: background, skin, left/right brow, left/right eye, upper/lower lip, left/right ear, nose, inner mouth, hair, hat, eyeglass, earring, necklace, neck, and cloth. The dataset is divided into $24,183$ images for the training, $2,993$ images for the validation, and $2,824$ images for the testing.

\textbf{LFW-PL}~\cite{lfwpl} contains  $2,927$ images of resolution $250\times 250$ with faces aligned in the centre. Face and hair regions are annotated using superpixel-based methods, thus resulting in inaccurate labels. The dataset is divided into $1,500$ images for the training, $500$ images for the validation, and $927$ images for the testing. 

\textbf{Helen}~\cite{Smith_2013_CVPR} is the most popular face parsing benchmark and it contains $2,330$ real-world images with rich context information. $11$ semantic labels are annotated: background, skin, left/right brow, left/right eye, upper/lower lip, inner mouth, nose and hair. There are significant annotation errors for facial skin and hair classes in the training set, as discussed in~\cite{Lin_2019_CVPR}, because these labels were automatically generated using image matting. The authors of Helen only cleaned the testing set to guarantee fair comparison in the test set. Helen dataset is divided into $2,000$ images for the training, $230$ images for the validation and $100$ images for the testing.

\textbf{LaPa}~\cite{lapa} is a face parsing dataset containing more than $22,176$ facial images with relatively more variations in expression, pose and occlusion. The same $11$ semantic classes are annotated as in Helen and the annotation process was guided by 106-point facial landmarks. The dataset is divided into $18,176$ images for the training, $2,000$ images for the validation, and $2,000$ images for the testing. The faces contain some variations in pose and occlusion but the background and hair region are largely removed since the faces are cropped with a hand-picked margin. 

\subsection{iBugMask: An In-the-wild Face Parsing Benchmark}
The proposed iBugMask consists of two parts: a training set obtained by pose augmentation and a manually curated testing set.  The training set contains $21,866$ images while the testing set contains $1,000$ images. We describe these two parts in detail below.
\subsubsection{A Large-Pose Augmented Training Set}
For machine learning models to learn to parse faces with large variations in head poses, the training set needs to contain a balanced distribution over poses. However, existing datasets contain faces mostly with absolute yaw angles less than 45 degrees. This means that models trained on these datasets cannot handle faces with extreme poses.

We propose to solve this problem by synthesising training faces with large poses. First, we examined the training set Helen, and manually corrected the labelling errors as in~\cite{Lin_2019_CVPR}. Next, we augmented the data with a face profiling method~\cite{3ddfa} that has been applied to augment face alignment datasets. One major advantage of this method is that it creates 3D meshes for both internal face and external face regions, which preserve the unpredictable hair regions as well as important context information for face parsing. Through face profiling, we augment the training set of Helen to a large scale one with many faces having large variation in head poses.
\begin{figure}[ht]
    \centering
    \includegraphics[width=\columnwidth]{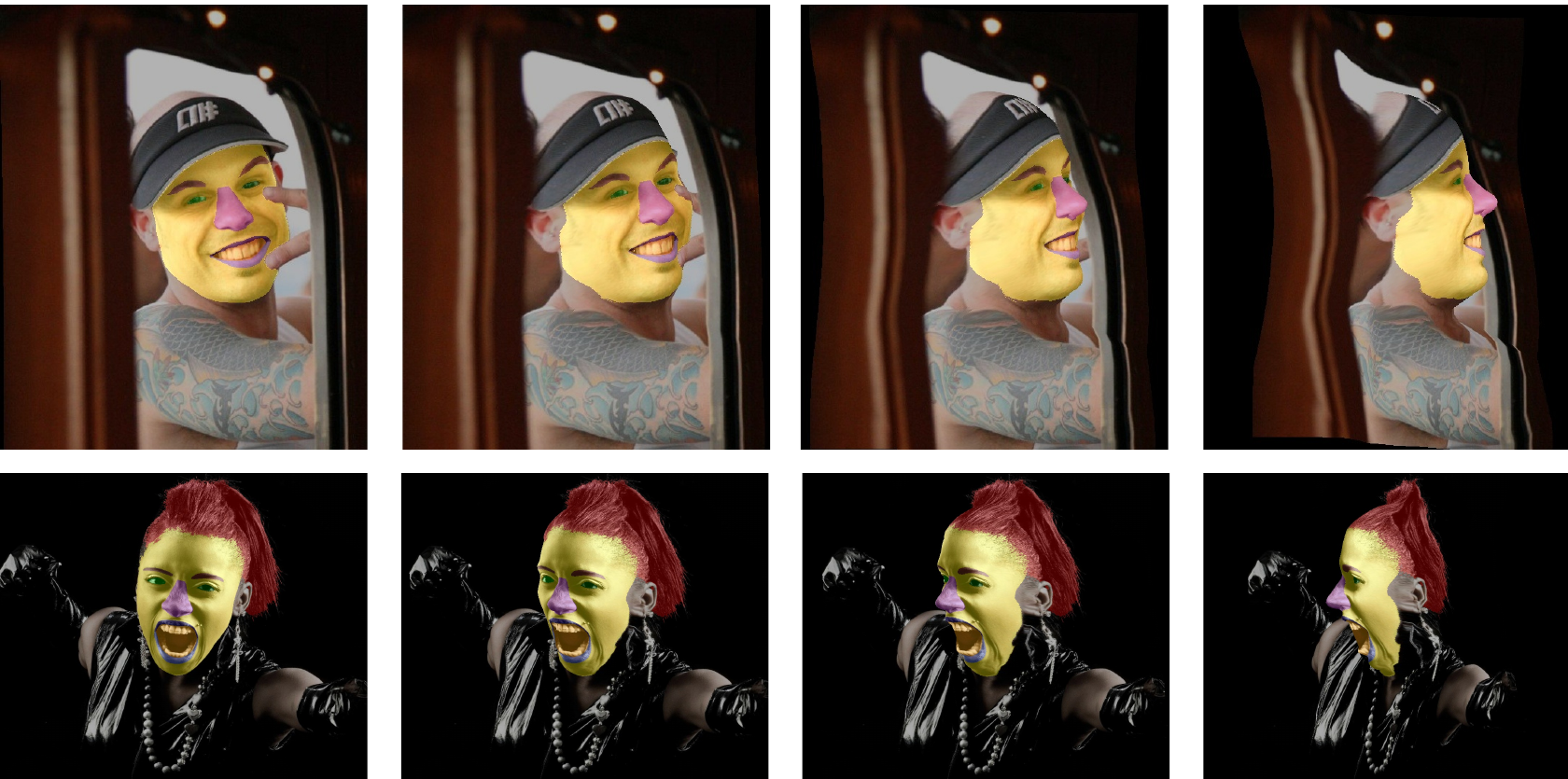}
    \caption{Examples of face data augmentation using 3DDFA~\cite{3ddfa}. The first column shows the original images and the other three columns show synthesised images with different $\Delta yaw$ until $yaw=90^{\circ}$.}
    \label{fig:helenlp-augmentation}
\end{figure}

With the fitted 3D model, we gradually enlarge the yaw angle of image at the step of $5^{\circ}$ until $90^{\circ}$.  Considering that the fidelity of a synthesised face is negatively related with the $\Delta yaw$, we resample the augmented images of each face with probabilities $0.8^{\Delta yaw/5^{\circ}}$. 
In \tablename~\ref{tab:helenlp-compare}, we compare our training set with other training sets and show that ours contains much more variations in pose, facial expression, and background. 
\begin{table}[h]
\begin{center}
\resizebox{\columnwidth}{!}{%
\begin{tabular}{l|c|c|c|c|c}
\Xhline{1.0pt}
\multirow{2}{*}{Benchmark} &
  \multirow{2}{*}{\begin{tabular}[c]{@{}c@{}}Number \\ of images\end{tabular}} & 
  \multirow{2}{*}{\begin{tabular}[c]{@{}c@{}}In-the-\\ wild\end{tabular}} &
  \multirow{2}{*}{\begin{tabular}[c]{@{}c@{}}Non-neutral \\ \& non-smile \end{tabular}}  &

  \multicolumn{2}{c}{$|yaw|$}  \\
          & &  &  & $\ge 30^{\circ}$ & $\ge 60^{\circ}$ \\ \hline
Helen~\cite{Smith_2013_CVPR}      & $2,000$  & \cmark     &   $40.1\%$    &    $120$              &      $4$             \\
CelebAMask-HQ~\cite{CelebAMask-HQ} & $27,176$  & \xmark & $44.6\%$  &   $1,565$           &       $60$          \\
LFW-PL~\cite{lfwpl}     & $2,000$ & \xmark  &  $74.7\%$     &  $125$            &     $0$             \\
LaPa~\cite{lapa}     & $18,176$ & \xmark  &  $39.7\%$     &  $3,961$            &     $194$             \\
iBugMask~(ours)   & $21,866$  & \cmark       & $34.6\%$    & $14,692$            &     $6,880$            \\
\Xhline{1.0pt}
\end{tabular}%
}\end{center}
\caption{Comparison of training sets. Ours has large variations in pose, expression and background.}
\label{tab:helenlp-compare}
\end{table}
We conduct extensive experiments in Sec.~\ref{sec:exp}. The results show that all models trained on our augmented training set improve over their counterparts trained on other datasets for in-the-wild face parsing.

\subsubsection{A Manually Curated Testing Set}
In in-the-wild images, faces can appear at any location in an image, with various distracting contextual information around it. In existing benchmarks, the target faces are cropped and centred by the data providers, largely removing background, part of hair and other faces. This introduces bias and evaluating methods on pre-processed images does not honestly reveal their robustness to the distracting context noise.

\begin{figure}[ht]
    \centering
    \includegraphics[width=0.8\linewidth]{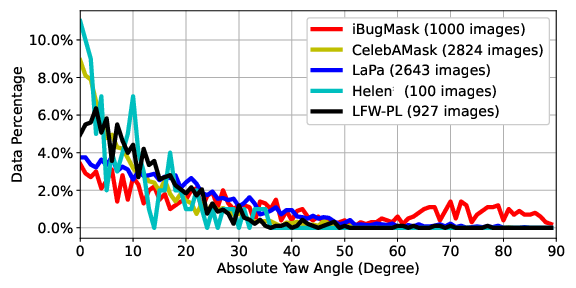}
    \caption{Absolute yaw angle distributions of different testing sets. Yaw is estimated with 3DDFA~\cite{3ddfa}.}
    \label{fig:ibugmask-yaw-dist}
\end{figure}

To fairly evaluate face parsing models under in-the-wild conditions, we present iBugMask dataset that contains $1,000$ challenging face images and manually-annotated labels for 11 semantic classes: background, facial skin, left/right brow, left/right eye, nose, upper/lower lip, inner mouth and hair. The images are curated from challenging in-the-wild face alignment datasets, including 300W~\cite{300w} and Menpo~\cite{deng2018menpo}. Compared to the existing face parsing datasets, iBugMask contains in-the-wild scenarios such as ``party'' and ``conference'', which include more challenging appearance variations or multiple faces. 
There is a larger number of profile faces. More expressions other than "neutral" and "smile" are also included (\eg~"surprise" and "scream"). Examples can be found in the rightmost column of \figurename~\ref{fig:benchmarks}.
\begin{figure}[tb]
    \begin{center}
    \subfloat[Pixel distribution for facial parts]{{\includegraphics[width=0.45\linewidth]{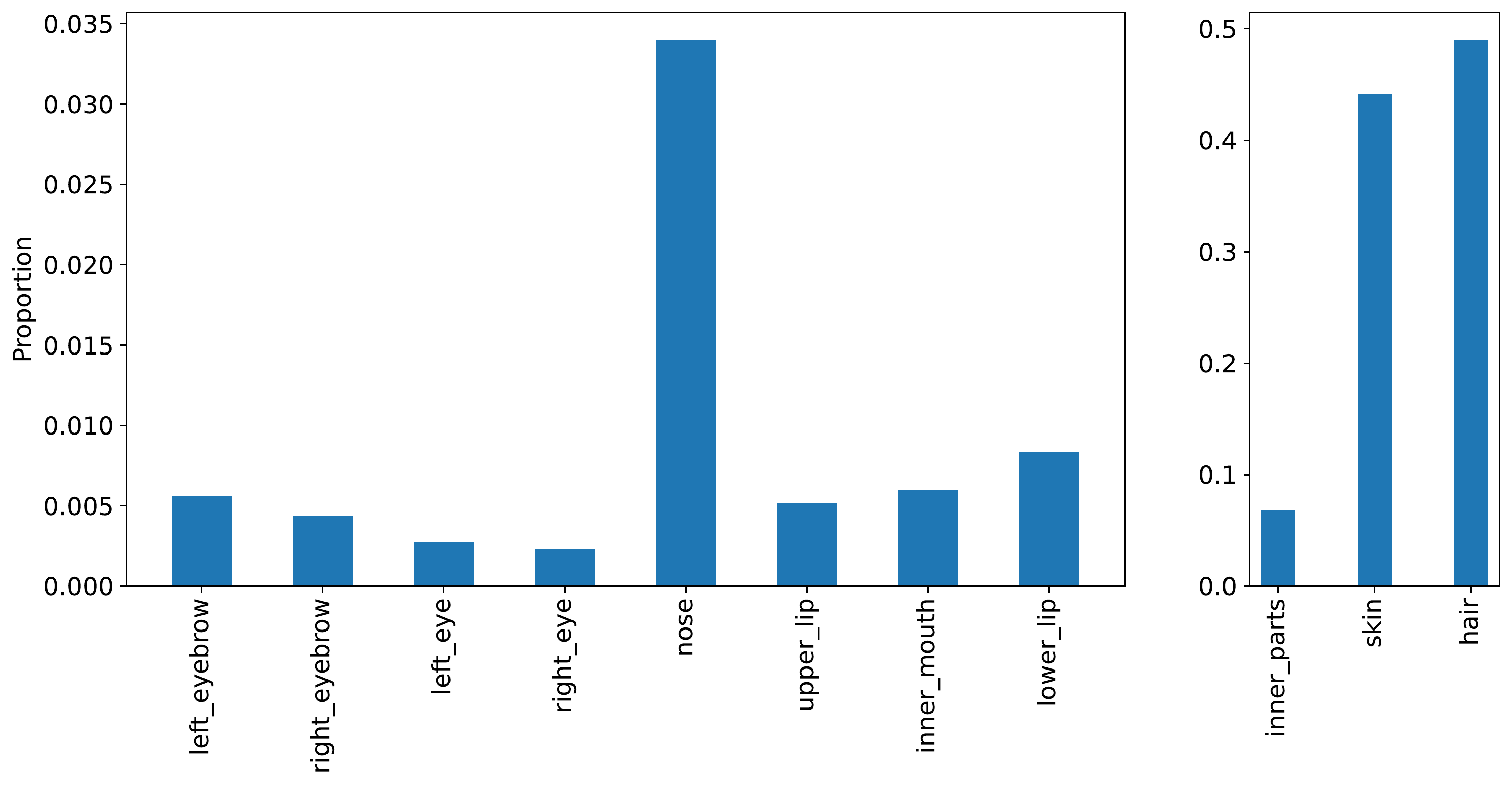} }} 
    \unskip\ \vrule\ 
    \subfloat[Region distribution for facial parts]{{\includegraphics[width=0.45\linewidth]{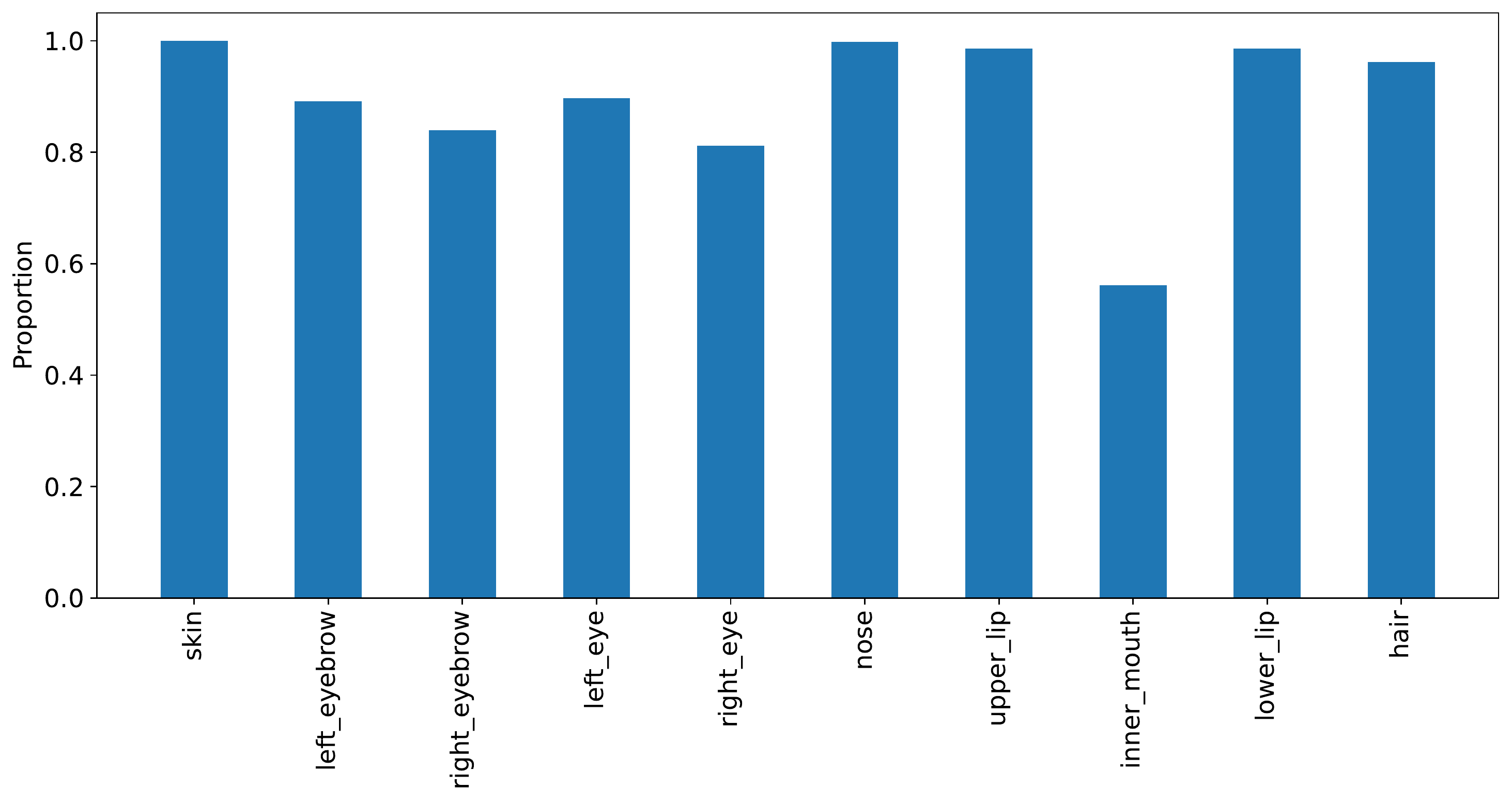} }}%
    \caption{Distributions for facial parts in iBugMask. Left: pixel distribution for facial parts. We compare different facial parts in the first subplot. We merge the inner parts to compare with skin and hair in the second subplot. Right: region distribution for facial parts. Inner mouth the least seen region in the dataset. }%
    \label{fig:pixel_dist}%
\end{center}

\end{figure}
\tablename~\ref{tab:ibugmask-compare} compares characteristics of different benchmarks. We use 3DDFA~\cite{3ddfa} to estimate the yaw angles with facial landmarks obtained by FAN~\cite{Bulat_2017_ICCV}. We use the facial expression classifier proposed by Want~\etal~\cite{Wang_2020_CVPR_suppressing} to estimate the facial expressions. \figurename~\ref{fig:ibugmask-yaw-dist} shows the absolute yaw angle distributions of benchmarks. Finally, \figurename~\ref{fig:pixel_dist} shows the pixel and region distributions in the testing set.
\begin{table}[h]
\begin{center}
\resizebox{\columnwidth}{!}{%
\begin{tabular}{l|c|c|c|c|c}
\Xhline{1.0pt}
\multirow{2}{*}{Benchmark} &
  \multirow{2}{*}{\begin{tabular}[c]{@{}c@{}}Number \\ of images\end{tabular}} & 
  \multirow{2}{*}{\begin{tabular}[c]{@{}c@{}}In-the-\\ wild\end{tabular}} &
  \multirow{2}{*}{\begin{tabular}[c]{@{}c@{}}Non-neutral \\ \& non-happy \end{tabular}}  &

  \multicolumn{2}{c}{$|yaw|$}  \\
          & &  &  & $\ge 30^{\circ}$ & $\ge 60^{\circ}$ \\ \hline
Helen~\cite{Smith_2013_CVPR}      & $100$  & \cmark     &   $40.0\%$    &    $4$              &      $0$             \\
CelebAMask-HQ~\cite{CelebAMask-HQ} & $2,824$  & \xmark & $42.4\%$  &   $180$           &       $6$          \\
LFW-PL~\cite{lfwpl}     & $927$ & \xmark  &  $73.3\%$     &  $53$            &     $7$             \\
LaPa~\cite{lapa}     & $927$ & \xmark  &  $40.8\%$     &  $525$            &     $48$             \\
iBugMask~(ours)   & $1,000$  & \cmark       & $56.1\%$    & $413$            &     $241$            \\
\Xhline{1.0pt}
\end{tabular}%
}\end{center}
\caption{Comparison of existing benchmark datasets. iBugMask has large variations in pose, expression and background.}
\label{tab:ibugmask-compare}
\end{table}
\section{Methodology}
\label{sec:method}
\begin{figure*}[h!]
    \centering
    \includegraphics[width=\linewidth]{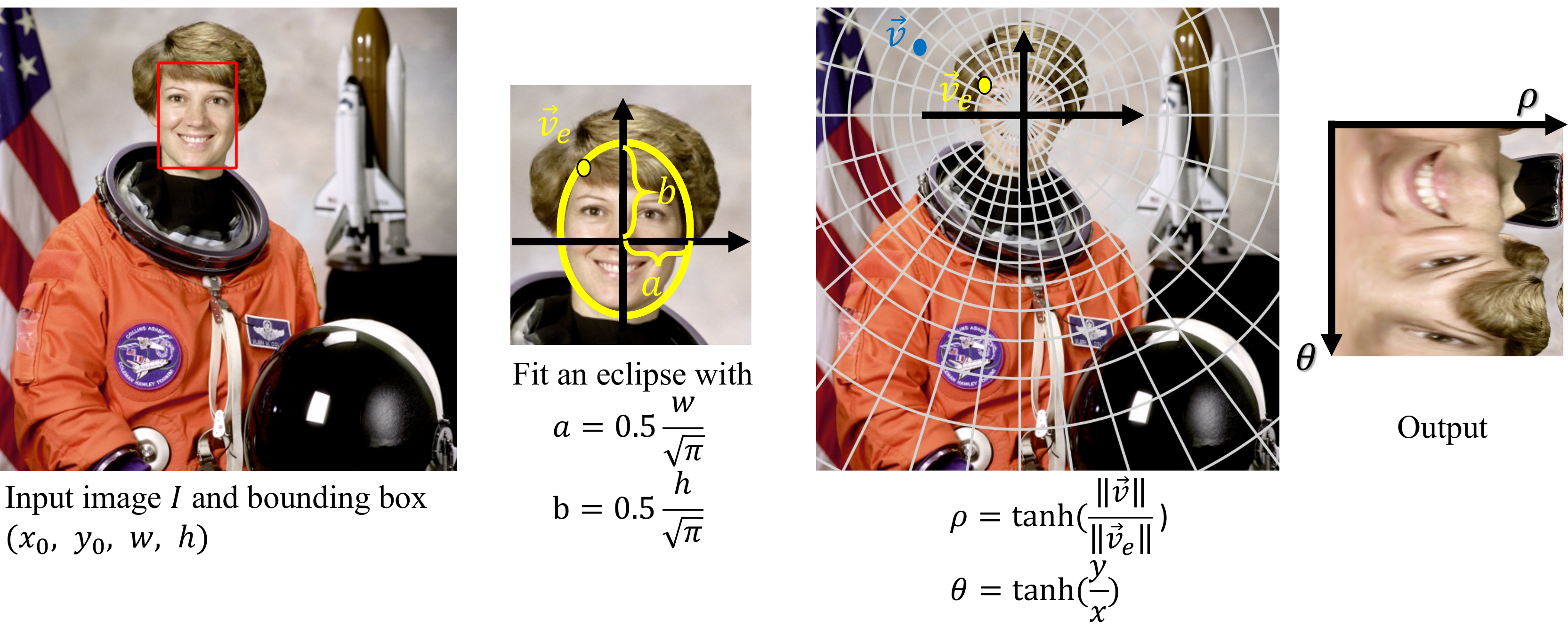}
    \caption{RoI Tanh-polar transform (RT-Transform). (1) Input is an image and the target bounding box. (2) An eclipse $e$ is fitted to the box. (3) The grey patterns depict the Tanh-polar sampling grid. $\Vec{v}$ is an arbitrary vector and  $\Vec{v}_e$ is on the eclipse $e$ in the same direction as $\Vec{v}$. (4) Transformed image. Due to the normalisation by $\Vec{v}_e$, the boundary of the face is located at $\rho=\tanh(1)=0.76$ regardless of the face size. All information is preserved and the proportion between face and background is fixed.}
    \label{fig:roi-tp-tsfm}
\end{figure*}
We introduce the RoI Tanh-polar Transformer Network for face parsing in the wild. \figurename~\ref{fig:overall} shows the overall framework: given an in-the-wild image in Cartesian coordinates and a bounding box of the target face, the whole image is first projected into the Tanh-polar space through the proposed RoI Tanh-Polar Transform in Section~\ref{sec:warping}. We further introduce a deep CNN encoder named HybridNet to extract semantic features of the Tanh-Polar-warped image. Consisting of several Hybrid Residual Representation Learning blocks (Section ~\ref{sec:HybridBlock}), the proposed HybridNet takes advantages of both Tanh-Cartesian and Tanh-Polar coordinate systems and thus can generate more robust spatial features. Those features are fed into a FCN decoder \cite{FCNs} to obtain Tanh-polar-based segmentation masks which are then mapped back into the Cartesian coordinate as the final output.
\subsection{RoI Tanh-Polar Transform}\label{sec:warping}
\subsubsection{To Crop or Not To Crop?} In in-the-wild face parsing, the target face is specified by a bounding box and is often not centralised. A common pre-processing step is to extend the facial bounding box with a certain margin and then to crop out the facial images, which are further resized into a certain resolution depending on the employed deep models. We refer to this pre-processing technique as \emph{crop-and-resize}. In this pre-processing approach, however, the cropping margin needs to be carefully selected. An overly loose margin may introduce irrelevant and distracting information, e.g. other faces, while a margin that are too narrow can lead to the ignorance of useful image regions like hairs, both of which are undesirable in the face parsing task. 
Another pre-processing method is to use facial landmarks for face alignment~\cite{liu2015faceattributes, Lin_2019_CVPR} such that the face is appropriately rotated. We refer to this method as \emph{align}. The landmarks can be jointly obtained with the face bounding boxes~\cite{retinaface}.

To overcome the limitations in the \emph{crop-and-resize} method and eliminate the need for facial landmarks, we propose the RoI Tanh-polar transform that warps the whole image to a canonical representation in the Tanh-polar space. Compared with the classical \emph{crop-and-resize} and \emph{align}, the only prerequisite of our method is the detected bounding box. Besides, our mapping can also introduce rotation equivariance to CNN models because of the polar-based representations.

The RT-Transform is illustrated in \figurename~\ref{fig:roi-tp-tsfm}. Let $\Vec{v}=(x,y)$ represent the Cartesian coordinate of a point in the original image, and let $w$ and $h$ represent the width and height of the bounding box, respectively. We select the centre of the bounding box as the polar origin. 
We first fit an ellipse $e$ to the target bounding box to the bounding rectangle, described by 
\begin{equation}
    \frac{x^2}{a^2}  + \frac{y^2}{b^2} =1, 
\end{equation}
where $a = 0.5 \frac{w}{\sqrt{\pi}}$, $b = 0.5\frac{h}{\sqrt{\pi}}$, and $w$ and $h$ are the width and height of the bounding box. 
We then define the Tanh-polar coordinate system by an injective map $f$: 
\begin{equation}\label{equ:tansform}
f (\Vec{v}) := (\tanh(\frac{y}{x}), \tanh(\frac{||\Vec v||}{||\Vec{v}_e||}))
\end{equation}
where $\Vec{v}_e={(x_e,y_e)}$ is the vector on the broader of the target face ellipse $e$ and $\Vec{v}_e$ and $\Vec{v}$ are \emph{parallel}. 
A new representation is constructed by resampling the input image over a rectangular grid in the Tanh-polar coordinate system. Following typical transformer networks~\cite{Jaderberg2015spatial, tai2019equivariant}, we use bilinear interpolation for points that do not coincide with the pixel locations in the input image. We name this as RoI Tanh-polar transform (RT-transform).
It can be observed that: 1) compared to representations obtained by crop-and-resize, all information in the input image is preserved in the new representation; 2) the normalisation with $\Vec{v}_e$ ensures that the target face always occupies around $76\%$ (since $\tanh(1)=0.76$). It is worth noting that the proposed RT-Transform is invertible and differentiable. Therefore, not only can the input RGB images be transformed but also the intermediate feature maps in CNNs. \\
\textbf{Rotation Equivariance.} To handle rotation of the target face, previous face parsing works~\cite{te2020edge,Lin_2019_CVPR} rely on transforming facial landmarks to canonical locations correct the rotation of the target face. We show that using the Tanh-polar representation can eliminate such pre-processing step.

The Tanh-polar coordinate system by definition is a canonical coordinate system~\cite{Segman1992canonical} for the \emph{rotation group} $\mathcal{SO}(2)$ with angle $\theta\in[-\pi,\pi]$. This is because for the rotation transformation $T_{\theta}\Vec{v}=( x\cos\theta- y\sin\theta,  x\sin\theta+ y\cos\theta)$, the Tanh-polar coordinate system satisfies~\cite{tai2019equivariant}
\begin{equation}\label{equ:cano}
    f(T_\theta \Vec{v}) = f(\Vec{v}) +  \mathbf{e}_{\theta},
\end{equation}
where $\mathbf{e}_\theta=(\theta,0)$. Thus, a rotation transformation $T_\theta$ appears as a translation by $(\theta,0)$ under the Tanh-polar coordinate system $f$. As a result, the planar convolution that is self-consistent with respect to translation~\cite{tai2019equivariant} in $f$ is now equivalent to $\mathcal{SO}(2)$ group-convolution~\cite{cohen2016group, esteves2018polar} in the Cartesian space. \\
\textbf{Scale Invariance.} Equation~\ref{equ:tansform} shows that the warped image would always have a fixed ratio between the face area and the background area regardless of the face's original size in the input images as long as the provided bounding box has the correct size. This means a model trained in the ROI Tanh-polar space would perform equally well on small faces as well as on large faces in the input image.

\begin{figure}[h]
    \centering
    \includegraphics[width=\linewidth]{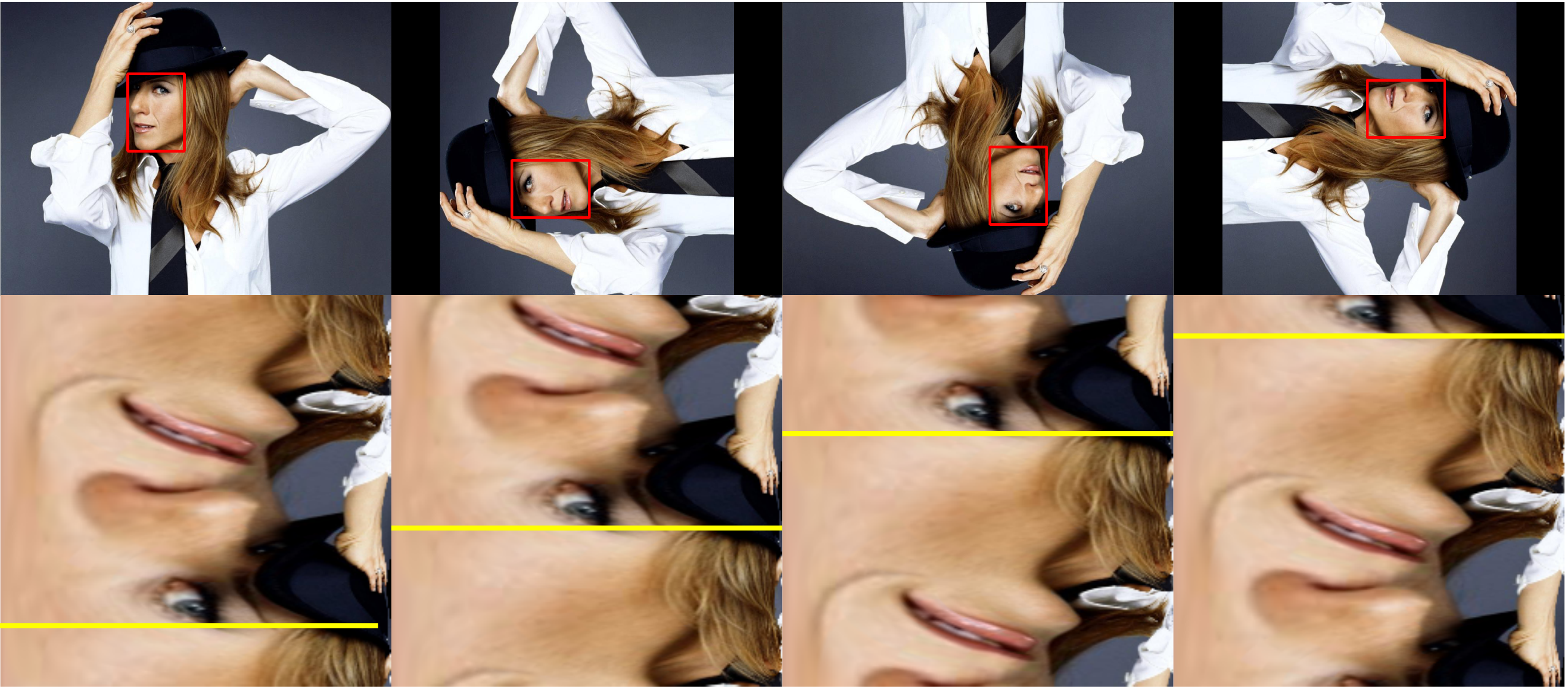}
    \caption{Rotation equivariance. Rotation is reduced to translation in the Tanh-polar coordinate system.}
    \label{fig:equivariance}
\end{figure}

\subsection{Hybrid Residual Representation Learning Block}\label{sec:HybridBlock}
Using the Tanh-polar representation as input to CNNs, rotation equivariance is achieved but translation equivariance may be lost. 
To overcome this, we propose Hybrid Residual Representation Learning Block, dubbed as \emph{HybridBlock}, a CNN building block similar to the Residual Block~\cite{resnet}.

The incentive of designing HybridBlock is to have two branches of convolutions learn representations that are complementary. One branch (Tanh-polar branch) learns the rotation equivariant representations while the other branch (Tanh-Cartesian branch) learns translation equivariant representations. The detailed components of a HybridBlock is depicted on the right in \figurename~\ref{fig:overall}.

We define Tanh-Cartesian coordinate system by 
\begin{equation}\label{equ:tanh-cart}
f_{TC} (\Vec{v}) := (\tanh(\frac{x}{||\Vec{v}_e||}), \tanh(\frac{y}{||\Vec{v}_e||})).
\end{equation}

The input to HybridBlock is a Tanh-polar representation $\mathbf{X}_{TP}$ of shape $(h,w,c)$. The residual path uses a stack of $1\times 1$, $3\times 3$ and $1\times 1$ convolutions following the \emph{bottleneck} design~\cite{resnet}. The first $1\times 1$ conv layer is used to reduce the channel dimension and its output feature maps are transformed to the Tanh-Cartesian space.
In each coordinate system a $3\times 3$ conv layer is used to compute feature maps, which are then concatenated in the Tanh-polar space. The last $1\times 1$ conv layer restores the channel dimension so the residual representation can be added to the input $\mathbf{X}_{TP}$. \\
\begin{figure}
    \centering
    \includegraphics[width=\columnwidth]{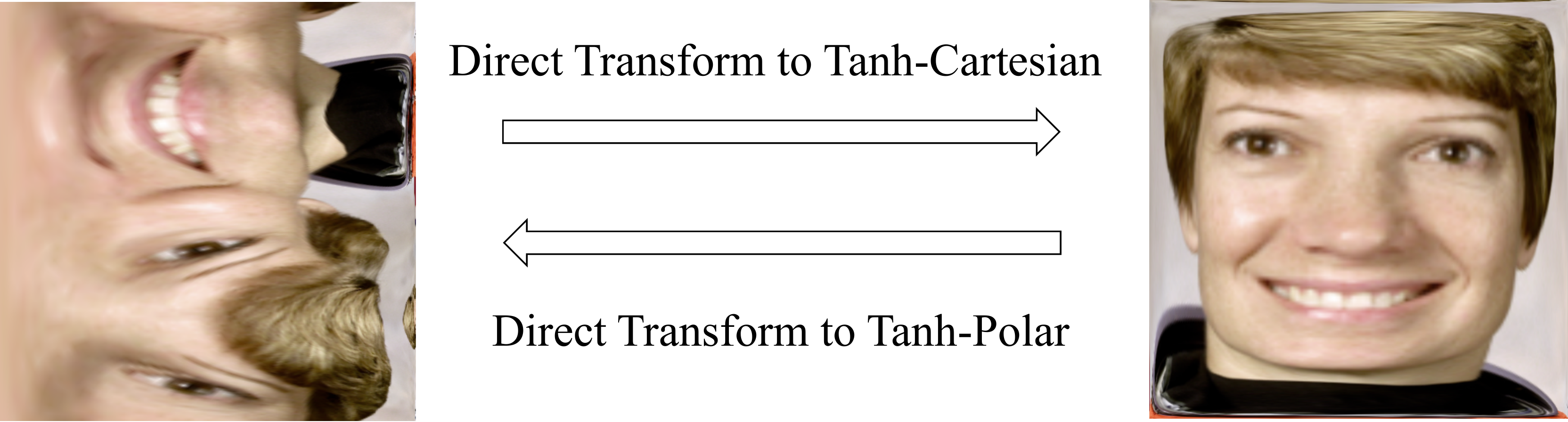}
    \caption{Direct transform between Tanh-polar and Tanh-Cartesian coordinates. We do not sample on the original image but directly transform between two coordinates. Translation equivariance is recovered in Tanh-Cartesian coordinates.}
    \label{fig:direct-tsfm}
\end{figure}
\textbf{Direct Transformation from Tanh-polar to Tanh-Cartesian.} To obtain Tanh-Cartesian representations, a naive approach is to inverse-transform from $f$ to Cartesian and then resample with Equ.~\ref{equ:tanh-cart}. However, iterated resampling will degrade image quality and amplify the influence of interpolation artefacts. To  circumvent this issue, we find the correspondence between the sampling grids in both coordinates and directly resample the Tanh-polar representation. \\
\textbf{Hybrid Receptive Field.} The receptive field~(RF) is the region in the input space that a particular neuron is looking at. The two $3\times3$ convolution layers in different coordinate systems have RFs of different shapes. The Tanh-polar one has the arc-shaped RF while the other has the rectangle-shaped RF.\\ 
\textbf{50-layer HybridNet.} We follow the design of ResNets~\cite{resnet} and stack HybridBlocks to create a new backbone network HybridNet50. Thanks to the grouped conv1x1 and the conv3x3 with halved channels, the overall number of parameters are less than the ResNet50 backbone (23.5 M versus 17.8M).
\subsection{RTNet: the Overall Framework }
\label{sec:rtnet-overall}With the previously introduced components, we now describe the overall framework of RoI Tanh-polar transformer network (RTNet) for face parsing in the wild. As in \figurename~\ref{fig:overall}, RTNet is based on the FCN framework~\cite{FCNs}. An input image $I$ of arbitrary resolution with the target bounding box is transformed to $I_{tp}$ in the Tanh-polar space. By default, the size of $I_{tp}$ is set to be $512\times 512$. Next, HybridNet-50 is used to extract features from $I_{tp}$, followed by a naive FCN decoder head to predict the segmentation mask in the Tanh-polar space. Finally, the segmentation mask is inverse-transformed to Cartesian as the final output that has the same resolution with the input image $I$. \\ 
\textbf{FCN Decoder.} We use the FCN-8s~\cite{FCNs} decoder to predict the masks. More advanced decoders like ASPP~\cite{deeplabv3plus} require dedicated hyperparameter-tuning that may largely affect the performance. 
The adopted decoder consists of two conv3x3 layers and a bilinear upsampling layer to map the feature maps to pixel-wise prediction logits. \\
\textbf{Loss function.} We use Cross-Entropy loss $loss_{CE}$ and Dice loss $loss_{dice}$~\cite{milletari2016vnet}. Two losses are jointly optimised with a factor of $\lambda$ and the overall loss $l$ is
\begin{equation}\label{equ:hybrid-loss}
l =\lambda loss_{CE} + (1-\lambda) loss_{dice}.
\end{equation} The losses are computed on the Tanh-polar coordinates since the outputs are of the same size and the computation can be batched and accelerated. \\
\textbf{Mixed padding.} Zero-padding is used in most CNNs to keep feature map size.This is not for the Tanh-polar representation, as it is periodic about the angular axis. We use wrap-around padding on the vertical dimension and replication padding on the horizontal dimension. \\
\textbf{Bounding box augmentation.} To improve robustness of RTNet, we augment the input bounding box during training time by adding a random shift and a random scaling. The augmentation can also be conducted during test time to the input image multiple times, and the inverse-transformed prediction masks can be averaged for smoother results. 
\section{Experiments}
\label{sec:exp}
\subsection{Experiment Setup}\label{sec:baseline}
\subsubsection{Baseline Methods}
We adopt the following criteria to select baseline methods. First, the model should be able to parse inner facial components as well as hair. Second, it is open-sourced and we are able to re-produce the reported performance by re-training the model from scratch. Third, the number of hyper-parameters has to be relatively small so that the performance does not rely on advanced training techniques. This allows that the same training setup can be applied and training can finish with reasonable computing resources and time.
The selected models include the classic models like FCN~\cite{FCNs}, as well as the advanced ones, such as Deeplabv3+~\cite{deeplabv3plus} and SPNet~\cite{spnet}. 
We collected their open-sourced codes and built an unified benchmarking codebase such that the same training and evaluation procedures are ensured. 

\subsubsection{Training and Evaluation}
\textbf{Data.} Each model is trained with four datasets, \ie~Helen~\cite{Smith_2013_CVPR}, CelebAMask-HQ~\cite{CelebAMask-HQ}, LaPa~\cite{lapa} and ours, 
among which Helen, LaPa and ours have the same set of labelling classes so the models trained on them can be evaluated directly. CelebAMask-HQ has more labelling classes and we assigned those additional classes to the background during training and evaluation. The target bounding box in each image is generated from the groundtruth mask to eliminate the bias from face detection. 
\\
\textbf{Evaluation.} 
We adopt two popular metrics, intersection over union (IoU), and F1 score (F1) for evaluation. We report the metrics for all foreground classes and their mean. The predicted masks are evaluated on the original image scale. For methods with crop-and-resize pre-processing, we resize the predicted masks to the size of the cropped image and then zero-pad it to match the original image resolution. For our RTNet, we apply inverse RT-Transform to the predicted masks. We did not employ other common evaluation techniques such as multi-scale, flipping or multi-cropping.

\subsubsection{Implementation Details}
We use PyTorch \cite{NEURIPS2019_9015} to implement all baselines and our methods. The backbone networks are pre-trained on ImageNet \cite{deng2009imagenet}. We use Stochastic Gradient Descent~(SGD) to optimise the losses. The initial learning rates are set to $0.01$ and the poly learning rate annealing schedule is adopted with $power = 0.9$. All methods are trained for $50$ epochs and early stopping is adopted if the mean IoU on the validation set stops growing for $15$ epochs.
For all methods, we apply random scaling in the range of $[0.5, 2.0]$, random horizontal flip and random brightness as data augmentation methods during training. 
For our methods, we transform the entire image to $512\times 512$ with our RT-Transform. Batch size is set to $4$ in all experiments. All training and evaluation are conducted on two RTX 2080 Ti GPUs. 
\subsection{Results on iBugMask}
We compare our model with different baselines that use \emph{align} as the input pre-processing method. The alignment templates are adopted from open-sourced ArcFace~\cite{deng2018arcface} library\footnote{\url{https://git.io/JOrvm}}. These templates have been shown successful in face recognition tasks. We report results for all facial parts. Eyebrows, eyes, lips and inner mouth are merged to Inner Parts.

\tablename~\ref{tab:ibugmask-benchmark} shows the benchmarking results. Our first observation is that iBugMask is challenging and cannot be readily solved. Compared with existing benchmarks, the models' performance on iBugMask is not saturated. For example, the mean F1 score on Helen has reached over $90\%$ but our best results on iBugMask are around $86\%$. We believe iBugMask can serve as a challenging benchmark for face parsing in the wild. 

Our second observation is that when using face alignment for pre-processing, the baseline models perform comparably on inner parts. However, the performance on hair is largely degraded because the templates cannot handle different hairstyles. In contrast, RT-transform allows our model to capture complete hair and face regions without being cut out. 

Lastly, without landmarks and alignment, our RTNet perform better than other methods in eyes, eyebrows, skin and hair regions, and comparably in nose, lips and mouth. When compared to the baseline FCN, we observe a large improvement in eyebrows and eyes. This could be attributed to the fact that the hybrid representation can better capture elongated regions.

\begin{table*}[]
\resizebox{\linewidth}{!}{
    \centering
    \begin{tabular}{c|cccccccc|c|c|c}
    \Xhline{1.0pt}
    & L-brow & R-brow & L-eye & R-eye & Nose & U-lip & I-mouth & L-lip 
    & Inner Parts & Skin & Hair   \\ \hline
    Deeplabv3~\cite{deeplab} & 70.6 & 69.5 & 78.6 & 78.4 & \textbf{90.2} & \textbf{75.7} & 82.2 & \textbf{78.6 }& \textbf{85.8} & 91.0 & 58.1  \\
    Deeplabv3+~\cite{deeplabv3plus} & 71.8 & 72.1 & 77.8 & 78.9 & 90.0 & 75.2  & 82.4 & 78.4  & \textbf{85.8} & 91.1 & 57.7 \\
    PSPNet~\cite{pspnet} & 70.2 & 70.0 & 78.6 & 79.1 & 89.5 & 75.3 & 82.2 & 78.1 & 85.3 & 90.7 & 58.2 \\
    SPNet~\cite{spnet} &73.2  & 71.9 & 77.9 & 78.0 & 90.0 & \textbf{75.7} & 81.7 & 78.5 & 85.5 & 90.1 & 57.9   \\ \hline
    FCN (baseline) & 71.0 & 70.6 & 78.2 & 78.5 & 89.6 & 75.6 & \textbf{82.7} & 78.0 & 85.4 & 90.9 & 57.7   \\
    RTNet~(ours) &  \textbf{76.0} & \textbf{73.0} & \textbf{79.6} & \textbf{79.9} & 89.3 & 75.5 & 82.5 & 77.6 & \textbf{85.8} & \textbf{91.8} & \textbf{81.8 }
    \\ \Xhline{1.0pt}
    \end{tabular}    

}
    \caption{Results on iBugMask. F1 scores are reported in percentage. Eyebrows, eyes, lips and inner mouth are merged to Inner Parts.}
    \label{tab:ibugmask-benchmark}
\end{table*}

\subsection{Qualitative Results}
\begin{figure*}
    \centering
    \includegraphics[width=\linewidth]{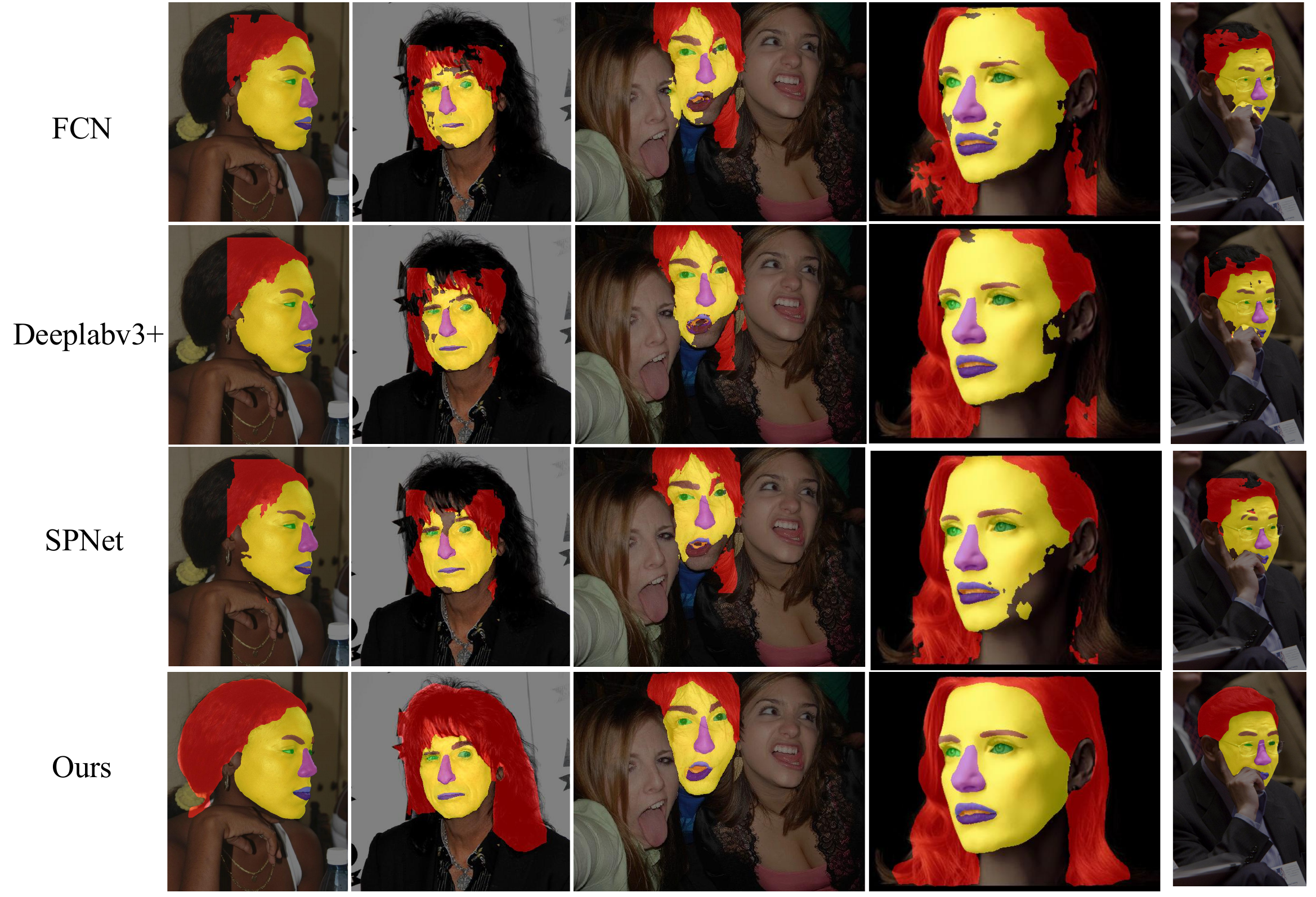}
    \caption{Qualitative results on iBugMask of four methods: FCN~\cite{FCNs}, Deeplabv3+~\cite{deeplabv3plus}, SPNet~\cite{spnet} and ours. Our method can handle large variations in head pose, hair styles, expressions and occlusions.
    }
    \label{fig:qualitative}
\end{figure*}

\figurename~\ref{fig:qualitative} visualises the prediction results of different methods, and our RTNet can better capture the varying hair styles, profile poses, occlusions, \etc, which again verifies the superior performance of our method under in-the-wild scenarios.

\subsection{Ablation Study}
We conduct extensive ablation studies to better understand the working mechanisms in RTNet.
All variants in this section were trained on pose-augmented images and evaluated on iBugMask.

\subsubsection{Effectiveness of RT-Transform}
We compare the performance of 4 pre-processing techniques: \\
1) \emph{Resize}: resizing all the input images to $512\times 512$ with zero-padding to preserve the aspect ratios; \\
2) \emph{Crop-and-resize}: cropping the face out with 40\% margin and then resizing the cropped face to $512\times 512$; \\
3) \emph{Align}: we use $5$ landmarks returned by RetinaFace~\cite{retinaface} to align the target faces using the open-source library and warp to $512\times 512$; \\
4) \emph{RT-Transform}: warping the whole image to a representation of size $512\times 512$ in the Tanh-polar space with the proposed RT-Transform. 

\tablename~\ref{tab:ablation-input} shows the F1 scores of different pre-processing methods on iBugMask. It can be seen that resizing the input images to the same size gives the lowest accuracy. This is in line with our expectation, as faces vary largely in size 
and uniformly resizing them will cause confusions. As for the crop-and-resize approach, only a small amount of improvement is observed, especially for the Hair class. 
This is potentially because
the pre-defined cropping margin cannot guarantee a full coverage of the hair region, which will cause accuracy loss on such regions. 
The alignment method gives competitive results on inner parts and facial skin. However, the performance on hair has degraded by a large amount because the warping template cannot account for different hair styles.
In contrast, our RT-Transform achieves the best performance on all three categories, and this can be attributed to the proposed Tanh-Polar transform that can emphasise the facial region while preserving all the contextual background information.
\begin{table}[!ht]
\begin{center}
    
\begin{tabular}{c|ccc}
\Xhline{1.0pt}
  Input Pre-processing&

  \begin{tabular}[c]{@{}c@{}}Inner \\ Parts \end{tabular} &
  \begin{tabular}[c]{@{}c@{}}Facial \\ Skin\end{tabular} &
  Hair  \\ \hline

Resize & 63.2 & 73.1 & 63.9  \\
\emph{Crop-and-resize}  &   79.0       &  81.3    &   71.3    \\
\emph{Align}  &   85.2       &  90.7   &   56.7    \\
RT-Transform & \textbf{85.8}  & \textbf{91.8} & \textbf{81.8}    \\
\Xhline{1.0pt}
\end{tabular}
\end{center}
\caption{Performance of our model with different pre-processing methods. F1 scores are reported in percentage.}
\label{tab:ablation-input}
\end{table}
\subsubsection{Design of HybridBlock}
We also conduct ablation studies to verify the design of HybridBlocks. We started with the Resnet50 backbone and gradually replace the residual blocks with HybridBlocks at different places of the network. In particular, the backbone comprises a stem layer and 4 stages of residual blocks, and we followed common practice~\cite{senet} to replace blocks in the last three stages with the proposed HybridBlock which has fewer parameters. Results of different replacing stages are reported in \tablename~\ref{tab:ablation-stages}, 
and we can observe that HybridBlocks can always introduce performance improvement with fewer parameters. Besides, the highest accuracy is achieved when using HybridBlocks in all the three stages, which demonstrate the effectiveness and the generality of the proposed HybirdBlocks.
\begin{table}[!h]
\begin{center}
    
\begin{tabular}{c|ccc|c}
\Xhline{1.0pt}
    \begin{tabular}[c]{@{}c@{}}Hybrid \\ Stages \end{tabular} &
    \begin{tabular}[c]{@{}c@{}}Inner \\ Parts \end{tabular} &
  \begin{tabular}[c]{@{}c@{}}Facial \\ Skin\end{tabular} &
  Hair & \# Params (in millions)
 \\ \hline
 -  &  85.6    & 90.7  & 81.2  & 31.4 \\
 Stage\_2 & 85.8 & 90.9 & 81.4 & 31.2 \\
 Stage\_3 & 85.6 & 91.6 & 81.2 & 29.9 \\
 Stage\_4 & 85.8 & 91.4 & 81.4 & 29.0 \\
 Stage\_All & \textbf{85.8} & \textbf{91.8} & \textbf{81.8} & \textbf{27.3} \\
\Xhline{1.0pt}
\end{tabular}
\end{center}
\caption{The effectiveness of HybridBlock in different stages of the backbone. F1 scores in percentage are reported.}
\label{tab:ablation-stages}
\end{table}

\subsubsection{Bounding box augmentation and mix-padding} 
\tablename~\ref{tab:ablation-augmentation} quantifies the performance gains of the bounding box augmentation and mix-padding described in Sec.~\ref{sec:rtnet-overall}. The box augmentation can make the model more robust to the bounding box noise. And mix-padding is necessary as the Tanh-polar representation is periodic about the angular axis. 
\begin{table}[ht]
\begin{center}
\begin{tabular}{cc|ccc}
\Xhline{1.0pt}
\begin{tabular}[c]{@{}c@{}}BBox \\ augment \end{tabular}  & Mix-padding &
  \begin{tabular}[c]{@{}c@{}}Inner \\ Parts \end{tabular} &
  \begin{tabular}[c]{@{}c@{}}Facial \\ Skin\end{tabular} &
  Hair 
  \\ \hline 
N & Y & 83.8 & 91.4 &  80.7  \\
Y & N & 85.0 & 91.4 & 80.8   \\
Y & Y & \textbf{85.8} & \textbf{91.8} & \textbf{81.8}             \\
\Xhline{1.0pt}
\end{tabular}
\end{center}
\caption{Ablation study. Random bounding box augmentation during training time, and mix-padding all contribute to improve the F1 scores (in percent).}
\label{tab:ablation-augmentation}
\end{table}

\subsection{Effectiveness of Pose-augmented Training Set}
To show the effectiveness of the pose augmentation, we train 6 on 4 different training sets. For simplicity and faster training, we use \emph{crop-and-resize} with $40\%$ margin to pre-process the input image to obtain a $512\times 512$ facial image for the baseline models.  We make the following observations:
\paragraph{Training on pose-augmented images improves all methods} We can also observe that training on pose-augmented images improved all methods, especially on the inner facial parts. This can be reasonably be attributed to that pose-augmented images is constructed in a way that the numbers of faces are balanced across different poses and that in-the-wild information is also preserved. In contrast, CelebAMask-HQ is a synthesised dataset with limited variations in pose and background. Although CelebAMask-HQ contains a larger number of facial images, models trained on this dataset achieve less competitive performance than trained on others.
\begin{table*}
\begin{center}
\resizebox{\linewidth}{!}{
\begin{tabular}{c|c|cccccc}
\Xhline{1.0pt}
Training Set                   & Region      & Deeplabv3~\cite{deeplab} & Deeplabv3+~\cite{deeplabv3plus} & FCN~\cite{FCNs}  & PSPNet~\cite{pspnet} & SPNet~\cite{spnet}  & RTNet (ours) \\ \hline
\multirow{3}{*}{Helen~\cite{Smith_2013_CVPR}}        & Inner Parts   & 70.3      & 71.7       & 68.1 & 70.4   & 69.8   & \textbf{74.3}         \\
                               & Hair    & 72.8      & 72.8       & 71.3 & 72.1   & 71.5   & \textbf{78.7}         \\
                               & Skin         & 90.7      & 90.5       & 88.7 & 90.2   & 89.6   & \textbf{91.9}         \\ \hline
\multirow{3}{*}{CelebAMask-HQ~\cite{CelebAMask-HQ}} & Inner Parts  & 73.6      & 73.7       & 73.9 & 74.0   & 74.4  & \textbf{76.1}         \\
                               & Hair         & 74.3      & 72.9       & 74.0 & 73.1   & 74.6   & \textbf{77.8}         \\
                               & Skin         & 88.8      & 88.6       & 89.1 & 88.6   & 89.7   & \textbf{91.8}         \\ \hline
\multirow{3}{*}{LaPa~\cite{lapa}}          & Inner Parts   & 74.2      & 74.1       & 75.1 & 74.4   & 75.9  & \textbf{77.6}         \\
                               & Hair         & 75.8      & 75.4       & 75.6 & 75.9   & 75.8   & \textbf{79.8}         \\
                               & Skin       & 89.8      & 89.3       & 90.1 & 89.7   & 89.9   & \textbf{92.2}         \\ \hline
\multirow{3}{*}{iBugMask-train~(ours)}       & Inner Parts  & 77.9      & 78.7       & 76.8 & 78.3   & 78.9 & \textbf{85.8}         \\
                               & Hair     & 72.4      & 72.7       & 64.6 & 72.0   & 72.9  & \textbf{81.8}         \\
                               & Skin     & 91.7      & 91.7       & 91.1 & 91.7   & 91.5  & \textbf{91.8}  \\
\Xhline{1.0pt}
\end{tabular}
}
\end{center}
\caption{Effectiveness of the pose-augmented training set iBugMask-train. Baseline models use \emph{crop-and-resize} for pre-processing. The mean F1 scores are reported (in percentage).}\label{tab:ibugmask-results}
\label{tab:bencmarking}
\end{table*}

\paragraph{RTNet consistently outperforms other methods}
The results on iBugMask show that our approach outperforms all other methods.
Moreover, when trained with our proposed pose-augmented images, RTNet significantly outperform all baselines,
especially on the hair class, which indicates that: 1) Compared with other benchmarking datasets, pose-augmented images can better benefit the in-the-wild learning of segmentation models, despite that most of its facial images were generated through the pose augmentation technique,
and 2) Different from the baselines, our RTNet can learn from the in-the-wild data more effectively and thus can demonstrate more robust performance on the unconstrained iBugMask dataset. 
\subsection{Comparison with the State-of-the-arts}
In additional to the self-collected iBugMask dataset, we also train and evaluate our method on various face parsing benchmarks. 

\textbf{Results on Helen.} \tablename~ \ref{tab:helenSOTA} compares our RTNet with other state-of-the-art methods on Helen~\cite{Smith_2013_CVPR,Lin_2019_CVPR}. Our model achieves slightly better performance on Facial Skin while significantly outperforms others on the Inner Parts and Hair classes.

\textbf{Results on LFW-PL.} 
\tablename~\ref{table:lfwpl-results} compares our RTNet with other state-of-the-art methods on LFW-PL~\cite{lfwpl}. Our model achieves comparable results in Inner Parts, and outperforms other methods in Facial Skin and Hair.

\begin{table}[h!]
\resizebox{\linewidth}{!}{
\begin{tabular}{l|ccc|c}
\Xhline{1.0pt}
Methods  & Skin & Hair & Background & accuracy \\
\hline
Liu \etal \cite{liu2015multi} & 93.93\% & 80.70\% & 97.10\% & 95.12\% \\
Long \etal \cite{FCNs} & 92.91\% & 82.69\% & 96.32\% & 94.13\% \\
Chen \etal \cite{deeplab} & 92.54\% & 80.14\% & 95.65\% & 93.44\% \\
Chen \etal \cite{chen2016semantic} & 91.17\% & 78.85\% & 94.95\% & 92.49\% \\
Zhou \etal \cite{zhou2017face} & 94.10\% & 85.16\% & 96.46\% & 95.28\% \\
Liu \etal \cite{Liu2017Face} & \textbf{97.55}\% & 83.43\% & 94.37\% & 95.46\% \\
Lin \etal \cite{Lin_2019_CVPR} & 95.77\% & 88.31\% & 98.26\% & 96.71\% \\
\hline
RTNet & 95.85\% & \textbf{90.08}\% & \textbf{98.55}\% & \textbf{97.11}\% \\
\Xhline{1.0pt}
\end{tabular}}
\caption{Comparison with state-of-the-art methods on LFW-PL. F1 scores for each region and the overall pixel accuracy are reported.}\label{table:lfwpl-results}

\end{table}
\textbf{Results on LaPa.} 
\tablename~\ref{tab:lapaSOTA} compares results from different methods on LaPa~\cite{lapa} dataset and we can easily spot that our RTNet consistently exhibits the highest performance on all three categories, which further demonstrates the generality of our method.

\subsection{Model Efficiency}
We also examine the running efficiency of different models by evaluating 1) the number of model parameters, 2) Floating Point Operations per Second (FLOPS) and 3) the actual inference time per sample. 
All models are measured on the same machine with a GTX1080Ti GPU with an $(512,512,3)$ input size. 
To ensure a fair comparison, we repeat the evaluation process 100 runs for each method and report the average. 
As shown in \tablename~\ref{tab:efficiency}, our model has the smallest model size and also operates with the fewest FLOPS when compared with three representative face parsing approaches. 
Although the inference time of our models are slightly longer than that of FCN and SPNet due to the direct sampling between two coordinates, we believe the time difference (16 ms) is tolerable as our method has shown improved performance over others.
\begin{table}[h!]
\resizebox{\linewidth}{!}{
\begin{tabular}{lcccc}
Measurement        & FCN         & Deeplabv3+ & SPNet & Ours                    \\ \Xhline{1.0pt}
Params (M)          & 32.95       & 39.64      & 39.13 & \textbf{27.29}          \\
FLOPS (GMac)        & 26.55       & 31.39      & 29.60 & \textbf{21.99} 
\\
Inference Time (ms) & \textbf{54} & 74         & 63    & 70 
\\ \Xhline{1.0pt} 
\end{tabular}
}
\caption{Efficiency comparison between four methods: FCN, Deeplabv3+, SPNet and ours. Input images are of size $(512,512,3)$. Models are profiled on the same machine and values are the mean of 100 runs. Lower values indicate better efficiency. Our model is more efficient in the number of parameters and FLOPS. M stands for Million, GMac for Giga Multiply–accumulate operations, ms for milliseconds.}\label{tab:efficiency}
\end{table}

\begin{table*}[t]
\resizebox{\linewidth}{!}{
    \centering
    \begin{tabular}{lccccccc|cccc}
    \Xhline{1.0pt}
    Methods &
      Eyes &
      Brows &
      Nose &
      I-mouth &
      U-lip &
      L-lip &
      Mouth &
      \begin{tabular}[c]{@{}c@{}}Inner\\  parts\end{tabular} &
      \begin{tabular}[c]{@{}c@{}}Facial \\ skin\end{tabular} &
      Hair &
      \begin{tabular}[c]{@{}c@{}}Foreground \\ mean\end{tabular}  \\ \hline
    Smith \etal \cite{Smith_2013_CVPR} & 78.5 & 72.2 & 92.2 & 71.3 & 65.1 & 70.0 & 85.7 & 80.4 & 88.2 & -    & - \\
    Zhou \etal \cite{icnet}            & 87.4 & 81.3 & 95.0 & 83.6 & 75.4 & 80.9 & 92.6 & 87.3 & -    & -    & - \\
    Liu \etal \cite{liu2015multi}      & 76.8 & 71.3 & 90.9 & 80.8 & 62.3 & 69.4 & 84.1 & 84.7 & 91.0 & -    & - \\
    Liu \etal \cite{Liu2017Face}       & 86.8 & 77.0 & 93.0 & 79.2 & 74.3 & 81.7 & 89.1 & 88.6 & 92.1 & -    & - \\
    Wei \etal \cite{wei2017learning}   & 84.7 & 78.6 & 93.7 & -    & -    & -    & 91.5 & 90.2 & 91.5 & -    & - \\
    Wei \etal \cite{Wei2019AccurateFI} & 89.0 & 82.6 & 95.2 & 86.7 & 80.0 & 86.4 & 93.6 & 91.5 & 91.5 & -    &   \\
    Lin \etal \cite{Lin_2019_CVPR}  & \textbf{89.6} & 83.1 & \textbf{95.6} & 86.7 & 79.6 & 89.8 & 95.0 & {92.4} & 94.5 & 83.5 & 88.6 \\  \hline
    RTNet &
      {89.3} &
      \textbf{84.9} &
      94.9 &
      \textbf{89.9} &
      \textbf{94.1} &
      \textbf{90.9} &
      \textbf{95.6} &
      \textbf{92.7} &
      \textbf{96.2} &
      \textbf{90.6} &
      \textbf{91.8} \\
    \Xhline{1.0pt}
    \end{tabular}
}
    \caption{Comparison with state-of-the-art methods on Helen. F1 scores are reported in percentage. \textbf{Bold} values are for the best results. }
    \label{tab:helenSOTA}
\end{table*}

\begin{table*}[t]
\resizebox{\linewidth}{!}{
    \centering
    \begin{tabular}{lccccccccccc}
   \Xhline{1.0pt} Methods & L-Eye & R-Eye & U-lip & I-mouth & L-lip & Nose & L-Brow & R-Brow & Skin & Hair & Mean \\ 
    \Xhline{1.0pt}
    Zhao~\etal~\cite{pspnet}    & 86.3  & 86.0  & 83.6  & 86.9    & 84.7  & 94.8 & 86.8   & 86.9   & 93.5 & 94.1 & 88.4 \\
    Liu~\etal~\cite{lapa}     & 88.1  & 88.0  & 84.4  & 87.6    & 85.7  & 95.5 & 87.7   & 87.6   & 97.2 & 96.3 & 89.8 \\
    Te~\etal~\cite{te2020edge}      & 89.5  & 90.0  & 88.1  & 90.0    & 89.0  & 97.1 & 86.5   & 87.0   & 97.3 & 96.2 & 91.1 \\ \hline
    Ours    & \textbf{91.5}  & \textbf{90.9}  & \textbf{88.7}  & \textbf{90.5}    & \textbf{90.5}  & \textbf{96.9} & \textbf{90.1}   & \textbf{89.1}   & \textbf{97.8} & \textbf{96.5} & \textbf{92.5}
   \\ \Xhline{1.0pt} \end{tabular}
   }
    \caption{Comparison with state-of-the-art methods on the LaPa benchmark. F1 scores are reported in percentage.}
    \label{tab:lapaSOTA}
\end{table*}
\section{Conclusion}
\label{sec:conclusion}
In this paper, we have approached in-the-wild face parsing from three aspects: data, representation and model. We have proposed a novel benchmark, iBugMask, for training and evaluating face parsing methods in unconstrained environment.  We have created a large-scale training set using pose augmentation and shown its effectiveness. We have solved the dilemma of face cropping and eliminated the need for facial landmarks by proposing a new Tanh-polar representation obtained by the proposed RoI Tanh-polar transform. Equivareriance with respect to rotations has also been achieved with the new representation. HybridBlock is introduced to extract features in both Tanh-polar and Tanh-Cartesian coordinates.
We have achieved the state-of-the-art performance on iBugMask as well as other existing face parsing benchmarks.
We expect our RT-Transform to be applicable to other face analysis tasks, where the heuristic pre-processing steps, such as cropping with bounding boxes and rotation correction with landmarks, are unavoidable.
\section*{Acknowledgements}
All datasets used in the experiments were obtained by, and all training, testing, and ablation studies have been conducted at, Department of Computing, Imperial College London, UK.

\bibliography{main}

\end{document}